\documentclass[a4paper,11pt]{elsarticle}

\usepackage{macrosArticle}

\usepackage[justification=centering]{caption}

\newcommand{\dmax}{d_{\max}}

\newcommand{\dmaxE}{\hat{d}_{\max}}

\newcommand{\MisC}{\mathrm{MisC}}
\newcommand{\Err}{\mathrm{Error}}

\title{A Spectral Algorithm with Additive Clustering for the Recovery of Overlapping Communities in Networks}
\author{Emilie Kaufmann$^{1}$, Thomas Bonald$^{2*}$ and Marc Lelarge$^{3}$\footnote[1]{Thomas Bonald and Marc Lelarge are members of the LINCS, Paris, France. See www.lincs.fr.} \\ 
\small CNRS \& CRIStAL, Univ. Lille$^1$, Telecom ParisTech$^2$, Inria \& Ecole Normale Sup\'erieure$^3$}

\begin{document} 

\selectlanguage{english}

\maketitle

\begin{abstract}
This paper presents a novel \emph{spectral algorithm with additive clustering}, designed to identify overlapping communities in networks.  
The algorithm is based on geometric properties of the spectrum of the expected adjacency matrix in a random graph model that we call \emph{stochastic blockmodel with overlap} (SBMO). An adaptive version of the algorithm, that does not require the knowledge of the number of hidden communities, is proved to be consistent under the SBMO when the degrees in the graph are (slightly more than) logarithmic. The algorithm is shown to perform well on simulated data and on real-world graphs with known overlapping communities. 
\end{abstract}

\section{Introduction}

Many  datasets (e.g., social networks, gene regulation networks)  take the form of graphs  whose structure depends on some underlying {\it communities}.  The commonly accepted definition of a community is that nodes tend to be more densely connected within a community  than with the rest of the graph. Communities are often hidden in practice and recovering the community structure directly from the graph  is a key step in the analysis of these datasets. Spectral algorithms are popular methods for detecting communities  \cite{VLux08Tuto}, that consist in two phases. First, a \emph{spectral embedding} is built, where the $n$ nodes of the graph are projected onto some low dimensional space generated by well-chosen  eigenvectors of some matrix related to the graph (e.g., the adjacency matrix or a Laplacian matrix). Then, a {\it clustering algorithm} (e.g., $k$-means or $k$-median) is applied to the  $n$ embedded vectors to  obtain a partition of the nodes into communities.

It turns out that the structure of many real datasets is better explained by {\it overlapping} communities. This is particularly true in social networks, in which the neighborhood of any given node is made of several social circles, that naturally overlap \cite{EgoSNAP}. 
Similarly, in co-authorship networks, authors often belong to several scientific communities and in protein-protein interaction networks, a given protein may belong to several protein complexes  \cite{Palla05Nature}. \ The communities do not form a partition of the graph and new algorithms need to be designed.
This paper presents a novel spectral algorithm, called spectral algorithm with additive clustering (SAAC). The algorithm consists in a spectral embedding based on the adjacency matrix of the  graph, coupled with an additive clustering phase designed to find overlapping communities. The proposed algorithm does not require the knowledge of the number of communities present in the network, and can thus be qualified as adaptive.

SAAC belongs to the family of model-based community detection methods, that are motivated by a random graph model depending on some underlying set of communities. In the non-overlapping case, spectral methods have been shown to perform well under the stochastic block model (SBM), introduced by Holland and Leinhardt \cite{Holland83}. 
Our algorithm  is inspired by the simplest possible extension of the SBM to overlapping communities, we refer to as the {stochastic blockmodel with overlaps} (SBMO). In the SBMO, each node is associated to a binary membership vector, indicating all the communities to which the node belongs.
We show that exploiting an additive structure in the SBMO leads to an efficient method for the identification of overlapping communities. To support this claim, we provide consistency guarantees when the graph is drawn under the SBMO, and we show that SAAC exhibit state-of-the-art performance on real datasets for which ground-truth communities are known.

The paper is structured as follows. In Section 2, we cast the problem of detecting overlapping communities into that of estimating a membership matrix in the SBMO model, introduced therein. In Section 3, we compare the SBMO with alternative random graph models proposed in the literature, and review the algorithms inspired by these models. In Section 4, we exhibit some properties of the spectrum of the adjacency matrix under SBMO, that motivate the new SAAC algorithm, introduced in Section 5, where we also formulate theoretical guarantees for an adaptive version of the algorithm. Section 6 illustrates the performance of SAAC on both real and simulated data and we discuss sparse SBMO in Section 7.

\paragraph{Notation} We denote by $||x||$ the Euclidean norm of a vector $x\in\R^d$. For any matrix $M\in\R^{n\times d}$, we let $M_{i}$ denote its $i$-th row and $M_{\cdot,j}$ its $j$-th column. For any $\cS\subset \{1,\ldots,d\}$, $|\cS|$ denotes its cardinality and $\ind_{\cS}\in \{0,1\}^{1\times d}$ is a row vector such that $(\ind_{\cS})_{1,i}=\ind_{\{i\in\cS\}}$. 
The Frobenius norm of a matrix $M\in\R^{n\times d}$ is 
\[||M||_F^2 = \sum_{i=1}^n ||M_{i}||^2 = \sum_{j=1}^d ||M_{\cdot,j}||^2=\sum_{1\leq i,j \leq n} M_{i,j}^2.\]
The spectral norm of a symmetric matrix $M\in\R^{d\times d}$ with eigenvalues $\lambda_1,\dots, \lambda_d$ is $||M||=\max_{i=1..d}|\lambda_i|.$
We denote by $\mathfrak{S}_K$ the set of permutation of $\{1,\dots,K\}$ and 
for $\sigma \in \mathfrak{S}_K$, by $P_\sigma\in\R^{K\times K}$ the permutation matrix associated to $\sigma$, defined by 
$(P_{\sigma})_{k,l} = \delta_{\sigma(k),l}.$

\section{The stochastic blockmodel with overlaps (SBMO)\label{sec:Models}}

\subsection{The model}

For any symmetric matrix $A\in [0,1]^{n\times n}$, let $\hat A$ be some random symmetric binary matrix whose entries $(\hat A_{i,j})_{i\le j}$ are independent Bernoulli random variables with respective parameters $(A_{i,j})_{i\le j}$. Then $\hat A$ is the adjacency matrix of an undirected random graph with expected 
adjacency matrix  $A$.
In all the paper, we  restrict the hat notation  to variables that depend on this random graph.
 For example,  the empirical degree  of node $i$ observed on the random graph and the expected degree of node $i$ are respectively denoted by
\[ \hat{d}_i = \sum_{j=1}^n \hat{A}_{i,j} \ \ \ \ \  \text{and} \ \ \ \ \ d_i = \sum_{j=1}^n A_{i,j}.\]
Similarly, we write $\hat{D}=\text{Diag}(\hat{d}_i)$,  $D=\text{Diag}(d_i)$, and
\[\dmaxE :=\max_{i}\sum_{j=1}^n{\hat{A}_{i,j}},\quad  \dmax = \max_{i}\sum_{j=1}^n{{A}_{i,j}}.\]

The stochastic block model (SBM) with $n$ nodes and $K$ communities
depends on some mapping $k:\{1,\ldots,n\} \rightarrow \{1,\ldots,K\}$ that associates
nodes to communities and on some symmetric community connectivity matrix
$B\in [0,1]^{K\times K}$. In this model, two nodes $i$ and $j$ are connected with probability  
\[{A}_{i,j} = B_{k(i),k(j)}=B_{k(j),k(i)}.\]
Introducing a membership matrix $Z\in \{0,1\}^{n\times K}$ such that $Z_{i,k}=\ind_{\{k(i)=k\}}$, the expected adjacency matrix can be written \[{A} = Z B Z^T.\]

The stochastic blockmodel with overlap (SBMO)  is a slight extension of
this model, in which $Z$ is only assumed to be in
$\{0,1\}^{n\times K}$  and $Z_i\ne 0$ for all $i$. 
  Compared to the SBM, the
 rows of the membership matrix $Z$ are no longer constrained to have only
one non-zero entry. Since these $n$ rows  give the communities of the respective $n$ nodes of the graph, this means that 
each node can now belong to several communities. Note that the SBMO yields implicit constraints on $B$ and $Z$, that should satisfy for all $i,j \in \{1,\dots,n\}$ that $Z_i BZ_j^T$ is smaller than 1.

\subsection{Performance metrics}

Given some adjacency matrix $\hat{A}$ drawn under the SBMO, our goal is to recover the underlying  communities, that is to build an estimate $\hat{Z}$ of the membership matrix $Z$, up to some permutation of its columns (corresponding to a permutation of the community labels). 
We denote by $\hat K$ the estimate of the number of communities ($K$ is in general unknown), so that $ \hat Z\in \{0,1\}^{n\times \hat K}$. 

We introduce two performance metrics for this problem. The first is related to the number of nodes that are ``well classified'', in the sense that there is no error in the estimate of their membership vector. 
The objective is to minimize the number of misclassified nodes of an estimate $\hat Z$ of $Z$, defined by 
  $\MisC(\hat{Z},Z)=n$ if
$\hat K\neq K$ and 
\begin{eqnarray*}
\MisC(\hat{Z},Z) = \min_{\sigma \in \mathfrak{S}_K}|\{i \in \{1,\ldots,n\}:\exists k\in \{1,\ldots,K\},
\hat{Z}_{i,\sigma(k)}\neq Z_{i,k}\}|
\end{eqnarray*}
otherwise.
The second  performance metric is  the fraction of wrong predictions  in the  membership matrix (again, up to a permutation of the community labels). We define the estimation error of $\hat{Z}$ as $\Err(\hat{Z},Z)=1$ if
$\hat K\neq K$ and otherwise by 
\[\Err(\hat{Z},Z) = \frac{1}{nK} \inf_{\sigma \in \mathfrak{S}_K} ||\hat{Z}P_\sigma - Z ||_F^2 \leq \frac{\MisC(\hat{Z},Z)}{n}.\]

\subsection{Identifiability}

The communities   of  a SBMO can only be recovered if  the model is {\it identifiable} in  that  the equality  $Z'B'{Z'}^T=ZBZ^T$, for some  integer $K'$ and matrices $Z'\in \{0,1\}^{n\times  K'}$, $B'\in [0,1]^{K'\times K'}$,    implies 
$\MisC(Z',Z)=0$ (and thus $K'=K$):  two SBMO with the same expected adjacency matrices have the same communities, up to a permutation of the  community labels. In this section, we derive sufficient  conditions for identifiability.

\begin{example}\label{ex:indent}
Consider the following SBMO with $n$ nodes and 3 overlapping communities:
\begin{equation}
B = \left( 
\begin{array}{ccc}
a&0&0\\
0&b&0\\
0&0&c
\end{array}
\right),\quad 
Z = \left(\begin{array}{ccc}
\mathbf{1}&\mathbf{1}&\mathbf{0}\\
\mathbf{0}&\mathbf{1}&\mathbf{1}\\
\mathbf{1}&\mathbf{0}&\mathbf{1}
\end{array} \right),
\label{OriginalSBM}
\end{equation}
where $a,b,c>0$ and $\mathbf{1}$ (resp. $\mathbf{0}$) is a vector of length $n/3$ with all
coordinates equal to $1$ (resp. $0$).
This SBMO is not identifiable since $ZBZ^T={Z'}{B'}{Z'}^T$ with
\[
{B'} = \left(\begin{array}{ccc}
a+b&b&a\\
b&b+c&c\\
a&c&a+c\end{array} \right), \quad {Z'} = \left(\begin{array}{ccc}
\mathbf{1}&\mathbf{0}&\mathbf{0}\\
\mathbf{0}&\mathbf{1}&\mathbf{0}\\
\mathbf{0}&\mathbf{0}&\mathbf{1}
\end{array} \right).
\]
Observe that this  is a  SBM with 3 non-overlapping communities.

\end{example}

In view of  the above example,  some additional assumptions are required to ensure identifiability. 
A first approach is to restrict the analysis to SBM. The following result is proved in \ref{proof:FormEigen}.

\begin{proposition}\label{prop:ID}
The SBMO is identifiable under the following assumptions:
\begin{itemize}
\item[] (SBM1) \ for all $\ell\neq k$, the rows $B_\ell$ and $B_k$ are different;
\item[] (SBM2) \ for all $i=1,\dots, n$, $\sum_{\ell=1}^K Z_{i,\ell}=1$.
\end{itemize}
\end{proposition}

Assumption (SBM1) is the usual condition for identifiability  of a SBM; the absence of overlap is enforced by assumption (SBM2).
Note that the SBM of Example \ref{ex:indent}  clearly
satisfies both assumptions and thus is identifiable:  this is the only SBM
 with expected adjacency  matrix $A=ZBZ^T$.
 One may wonder whether the SBMO is identifiable if 
 we impose an overlap, that is the existence of some node  $i$
such that $\sum_{\ell=1}^K Z_{i,\ell}\geq 2$. The answer is negative,
as shown by the following example.

\paragraph{Example \ref{ex:indent} (continued)}
{\it Without loss of generality, we assume that $c\leq \min(a,b)$.
Consider the following SBMO with $n$ nodes and 4 overlapping communities:
\[
{B''} = \left(\begin{array}{cccc}
a+b-c&b-c&a-c&0\\
b-c&b&0&0\\
a-c&0&a&0\\
0&0&0&c
\end{array}
\right),\quad 
{Z''} = \left(
\begin{array}{cccc}
\mathbf{1}&\mathbf{0}&\mathbf{0}&\mathbf{1}\\
\mathbf{0}&\mathbf{1}&\mathbf{0}&\mathbf{1}\\
\mathbf{0}&\mathbf{0}&\mathbf{1}&\mathbf{1}
\end{array}
\right).
\]
We have $ZBZ^T={Z''}{B''}{Z''}^T$.
}

\

Thus some  
additional assumptions are required 
to make the SBMO  identifiable. 
It is in fact sufficient that the community connectivity matrix  is invertible and that each
community contains at least one pure node (that is, belonging to this community only).
The following result is proved in \ref{proof:FormEigen}.

\begin{theorem}\label{thm:ID}
The SBMO is identifiable under the following assumptions:
\begin{itemize}
\item[](SBMO1) \ $B$ is invertible;
\item[] (SBMO2) \ for each $k=1,\dots,K,$ there exists $i$ such that $Z_{i,k}=\sum_{\ell=1}^KZ_{i,\ell}=1$,
\end{itemize}
\end{theorem}

Observe that the two SBMO of Example  \ref{ex:indent}, with membership matrices $Z$ and $Z''$, violate (SBMO2).
Only the SBM is   identifiable.
In particular,  if we generate a SBMO with 3
overlapping communities based on the matrices $B$ and $Z$, our algorithm 
will return at best 3 non-overlapping communities corresponding to the SBM
with membership matrix ${Z'}$.
To recover the model \eqref{OriginalSBM}, some additional information is required on the 
community structure. 
For instance, one may impose $K=3$ and that 
each node belongs to exactly two communities. Note that this last condition alone is not sufficient, in view of the third model of Example  \ref{ex:indent}.

Our choice for SBMO1-2 is motivated by applications to social networks: homophily will make the matrix $B$ diagonally dominant, hence invertible.
In the rest of the paper, we assume that the identifiability conditions (SBMO1) and (SBMO2) are satisfied.

\subsection{Subcommunity detection}
\label{sub:sub}
Any SBMO with $K$ overlapping communities may be viewed as a 
SBM with up to $2^K$ non overlapping
communities, corresponding to  groups of nodes sharing exactly the same communities in the SBMO and that we refer to as {\it subcommunities}.

Let ${K'}$ be the number of subcommunities in the SBMO:
\[{K'} = \left|\cT\right|, \ \ \text{where} \ \ \cT = \{ z \in \{0,1\}^{1\times K} : \exists i\in \{1,\ldots,n\}: Z_i = z\}.\]
The corresponding SBM  has ${K'}$ communities indexed by $z \in \cT$, 
with  community connectivity matrix ${B'}$ given by   
${B'}_{y,z} = y B z^T$ for all $(y,z) \in \cT^2$. 
The SBM of Example \ref{ex:indent}
can be derived from the first SBMO  in this  way for instance.
More interestingly, it is easy to check that if the initial SBMO
satisfies (SBMO1)-(SBMO2) then the corresponding SBM satisfies (SBM1)-(SBM2). 

\begin{figure}[ht]
\centering
\includegraphics[angle=-90,width=0.4\linewidth]{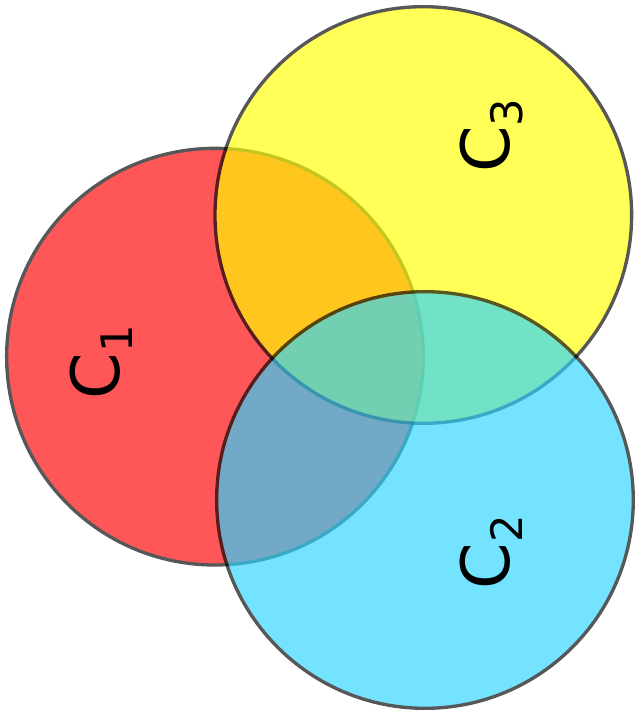}
\hspace{0.3cm}
\includegraphics[angle=-90,width=0.4\linewidth]{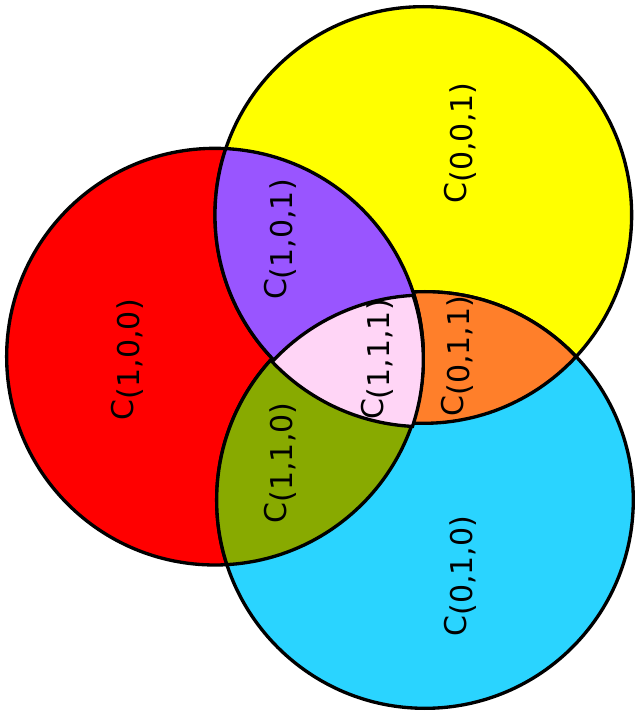}
\caption{Three overlapping communities of a SBMO (left) and the subcommunities of the associated SBM (right).}
\end{figure}

This suggests that 
  community detection in the SBMO reduces to community detection in the corresponding SBM,  for which many efficient algorithms are known. 
However, the notion of performance for a SBM is different from the that for the underlying SBMO: the knowledge of the subcommunities is not sufficient to recover the initial overlapping communities, that is to obtain an estimate $\hat{Z}$ such that $\MisC(\hat{Z},Z)$ is small. It is indeed necessary to map these subcommunities to elements of $\{0,1\}^K$, which is not an easy task: first, the number of communities $K$ is unknown; second, assuming $K$ is known, there are up to $2^K!$ such mappings so that a simple  approach by enumeration is not feasible in general. 
Moreover, the performance of clustering algorithms degrades rapidly with the number of communities so that it is preferable to work directly on the $K$ overlapping communities rather than on the $K'$  subcommunities, with $K'$ possibly as large as $2^K$.

Our algorithm detects directly  the $K$  overlapping communities using   the specific  geometry of the  eigenvectors of the  expected adjacency matrix, $A$.
We provide conditions under which  these geometric properties  hold for the observed adjacency matrix, $\hat A$, 
which guarantees the consistency of our algorithm: the $K$ communities are recovered with probability tending to 1 in the limit of a large  number of nodes $n$.

\newpage

\subsection{Scaling}\label{sec:Scaling}

To study the performance of our algorithm when the number of nodes $n$ grows, we  introduce a degree parameter $\alpha_n$ so that the expected adjacency matrix of a graph with $n$ nodes  is in fact given by
\[{A} = \frac{\alpha_n}{n}Z B Z^T,\]
with $B\in[0,1]^{K\times K}$ independent of $n$ and $Z\in
\{0,1\}^{n\times K}$.
Although $Z$ depends on $n$, we do not make it explicit in the
notation. Observe that the expected degree of each node grows like $\alpha_n$, since 
\[d_i=\alpha_n\left(\frac{1}{n}Z_i B Z^T{\bf 1}\right),\]
where  $ {\bf 1}$ is the vector of one's of dimension $n$. 

We assume that the set of subcommunities $\cT$ does not depend on $n$ and
that for all $z\in\cT$, there exists a positive constant (independent of $n$) $\beta_z$ such that: 
\begin{equation}
\quad \frac{|\left\{i :  Z_{i} = z\right\}|}{n} \to\beta_z.\label{def:BetaL}
\end{equation}
This implies the existence of positive constants $L_z$ and of a matrix $O\in\R^{K\times K}$, such that
\begin{equation}\forall z\in\cT, \ \frac{1}{n}zBZ^T {\bf 1} \to  L_z, \ \ \ \ \ \text{and} \ \ \ \ \ \frac{1}{n} Z^TZ \to O.\label{def:Lz}\end{equation}
One has $d_i\sim \alpha_n L_z$ for any $i$ such that  $Z_i=z$. In the sequel, we assume that the graph is sparse in the sense that $\alpha_n\to \infty$ with $\alpha_n/n\to 0$. Note that with previous assumptions, this condition ensures that the coefficients of the expected adjacency matrix $A$ are smaller than $1$.
Observe also that $O_{k,k}$ is the (limit) proportion of nodes that belong to community $k$ while $O_{k,l}$ is the (limit) proportion of nodes that belong to  communities $k$ and $l$, for any $k\ne l$. 
Hence we refer to $O$ as the \emph{overlap matrix}. 

In the following, we will slightly abuse notation by writing
$O=\frac{1}{n} Z^TZ$ and $d_i=\alpha_n L_z$ if $Z_i=z$, although these equalities in fact hold only in the limit.

\section{Related work\label{subsec:biblio}}

\paragraph{Models} 
Several random graph models have been proposed in the literature to model networks with overlapping communities. In these models, each node $i$ is characterized by some community membership  vector $Z_i$ that is not always a binary vector, as in the SBMO. In the Mixed-Membership Stochastic Blockmodel (MMSB) \cite{Airoldi08}, introduced as the first model with overlaps, membership vectors are probability vectors drawn from a Dirichlet distribution. In this model, conditionally to $Z_i $ and $Z_j$, the probability that nodes $i$ and $j$ are connected is $Z_iBZ_j^T$ for some community connectivity matrix $B$, just like in SBMO. However, the fact that $Z_i$ and $Z_j$ are probability vectors makes the model less interpretable. In particular, the probability that two nodes nodes are connected does not necessarily increase with the number of communities that they have in common, as pointed out by Yang and Leskovec \cite{YangLeskovec12AGM}, which contradicts a tendency empirically observed in social networks.

A first model that relies on binary membership vectors is the Overlapping Stochastic Block Model (OSBM) \cite{OSBM11}, in which two nodes $i,j$ are connected with probability  
$\sigma(Z_iWZ_j^T + Z_iV + Z_jU + w)$,
where $W\in \R_+^{K\times K}$, $U,V\in \R_+^K$, $w\in\R$, and $\sigma$ 
is the sigmoid function. Now
the probability of connectivity of two nodes  increases with the number of communities shared, but the particular form of the probability of connection makes the model hard to analyze. 
Given a community connectivity matrix $B$, another natural way to build a random graph model based on binary membership vectors is to assume that two nodes $i$ and $j$ are connected if any pair of communities $k,l$ to which these nodes respectively belong can explain the connection. In other words, $i$ and $j$ are connected with probability
$ 
 1 - \prod_{k,l=1}^K(1-B_{k,l})^{Z_{i,k} Z_{j,l}}.
$
Denoting by $Q$ the matrix with entries $Q_{k,l}=-\log(1-B_{k,l})$, this probability can be written
$
1 - \exp\left(-Z_i Q Z_j^T\right)\simeq  Z_i Q Z_j^T,
$
where the  approximation is  valid for sparse networks. In this case, the model is very close to the SBMO, with 
connectivity matrix $Q$. 
The Community-Affiliation Graph Model (AGM) \cite{YangLeskovec12AGM} is a particular case of this model in which $B$ is diagonal.
The SBMO with a diagonal connectivity matrix can be viewed as a particular instance of an Additive Clustering model \cite{ShepardArabie79} 
and is also related to the `colored edges' model \cite{Balletal11AGM}, in which $\hat{A}_{i,j}$ is drawn from a Poisson distribution with mean $\theta_i\theta_j^T,$ where $\theta_i \in \R^{1,K}$ is the (non-binary) membership vector of node $i$. Letting $\theta_{i}=\sqrt{B_{i,i}}Z_{i}$ and approximating the Poisson distribution by a Bernoulli distribution, we recover the SBMO.

The Overlapping Continuous Community Assignment Model (OCCAM), proposed by Zhang et al. \cite{Levina14Overlap} relies on overlapping communities but also on individual degree parameters, which generalizes the degree-corrected stochastic blockmodel \cite{KarrerNewman11}. In the OCCAM,  a  degree parameter $\theta_i$ is associated to each node $i$.  Letting $\Theta=\text{Diag}(\theta_i) \in \R^{n\times n}$, the expected adjacency matrix is 
$A = \Theta Z B Z^T \Theta$,
with a membership matrix $Z \in \R^{n\times K}$. Identifiability of the model is proved assuming that $B$ is positive definite, each row $Z_i$ satisfies $||Z_i||=1$, and the degree parameters satisfy $n^{-1}\sum_{i=1}^n \theta_i = 1$.  The SBMO can be viewed as a particular instance of the OCCAM, for which we provide new identifiability conditions, that allow for binary membership vectors. Recently, \cite{Panov17Overlap} proposed other identifiability conditions under OCCAM, still without degree-correction ($\Theta = \mathrm{Id}$) and for $Z_i$ that are probability vectors.

\paragraph{Algorithms} Several algorithmic methods have been proposed to identify overlapping community structure in networks \cite{Xie12SOA}. Among the model-based methods, that rely on the assumption that the observed network is drawn under a random graph model, some are approximations of the maximum likelihood or maximum a posteriori estimate of the membership vectors under one of the random graph models discussed above. For example, under the MMSB or the OSBM the membership vectors are assumed to be drawn from a probability (prior) distribution, and  variational EM algorithms are proposed to approximate the posterior distributions \cite{Airoldi08,OSBM11}. However, there is no proof of consistency of the proposed  algorithms. In the MMSB, \cite{Anand13MMSBT} propose the first consistency results, for an estimator based on the moment method. It is also referred to as a spectral method, as it requires to compute eigenvectors of a tensor, which is performed in practice using tensor power iteration. In this paper, spectral algorithms refer to simpler algorithms, that only require to compute the leading eigenvectors of a matrix associated to the graph.

The first occurrence of a spectral algorithm to find overlapping communities goes back to \cite{FuzzyCMeans07}. The proposed method is an adaptation of spectral clustering with the normalized Laplacian (see e.g., \cite{Newman13}) with a fuzzy clustering algorithm in place of $k$-means, and its justification is rather heuristic. Another spectral algorithm has been proposed by \cite{Levina14Overlap}, as an estimation procedure for the (non-binary) membership matrix under the OCCAM. The spectral embedding is a row-normalized version of $\hat{U} \hat{\Lambda}^{1/2}\in\R^{n\times K}$, with $\hat{\Lambda}$ the diagonal matrix containing $K$ leading eigenvalues of $\hat{A}$ and $\hat{U}$ the matrix of associated eigenvectors. The centroids obtained by a $k$-median clustering algorithm are then used to estimate $Z$. This algorithm is proved to be consistent under the OCCAM, when moreover degree parameters and membership vectors are drawn according to 
some distributions. Similar assumptions have appeared before in the proof of consistency of some community detection algorithms in the SBM or DC-SBM \cite{Zhaoetal12ConsDCSBM}. Our  consistency results are established for fixed parameters of the model, and hold for relatively sparse graph ($\alpha_n\simeq \log n$), unlike those obtained under the OCCAM.

\section{Spectral analysis of the adjacency matrix in the SBMO\label{sec:Heuristic}}

In this section, we describe the  spectral
structure of the  adjacency matrix in the SBMO.

\vspace{-0.2cm}

\subsection{Expected adjacency matrix\label{subsec:heuristic}}

Let $\cZ$ be the set of membership matrices that contains at least one pure node per community:
\[
\cZ =\{ Z\in \{0,1\}^{n\times K},\:\forall k=1,\ldots,K,\:\exists i\in
\{1,\ldots,n\},\: Z_{i,k}=\sum_{\ell}Z_{i,\ell}=1\}.
\vspace{-0.2cm}
\]
From the identifiability conditions  (SBMO1) and (SBMO2), $A=ZBZ^T$ is of rank $K$ (refer to the proof of Theorem ~\ref{thm:ID}) and $Z$ belongs to $\cZ$. 
Let $U\in\R^{n\times K}$ be a matrix whose columns $u_1,\dots,u_K\in \R^n$ are normalized orthogonal eigenvectors associated to the $K$ non-zero eigenvalues of $A$. 
The structure of $U$ is described in the following proposition. Its first statement follows from the fact that the eigenvectors $u_1,\dots,u_K$ form a basis of $\text{Im}(A)$ and that $\text{Im}(A)\subseteq \text{Im}(Z)$. Its second statement is established in the proof of  Theorem ~\ref{thm:ID}. 

\begin{proposition}\label{prop:Insight} 

\begin{enumerate}
 \item There exists $X\in\R^{K\times K}$ such that $U=ZX$.
 \item If $U=Z'X'$ for some $Z'\in \cZ$, $X'\in \R^{K\times K}$, then there exists $\sigma \in \mathfrak{S}_K$ such that $Z=Z'P_\sigma$.
\end{enumerate}
\end{proposition}

This decomposition reveals in particular an \emph{additive structure} in $U$: each row $U_i$ is the sum of rows corresponding to pure nodes associated to the communities to which node $i$ belongs. Fixing for each $k$ a pure node $i_k$ in community $k$, one has indeed
\begin{equation}\forall i, \ U_i = \sum_{k=1}^K U_{i_k}\ind_{(Z_{i,k}=1)}\label{AdditiveStructure}\end{equation}
Proposition \ref{prop:FormEigen}, proved  in~\ref{proof:FormEigen}, relates the eigenvectors of $A$ to those of a $K\times K$ matrix featuring the  overlap matrix $O$ introduced in Section~\ref{sec:Scaling}. Note that for any
$x\in \R^K$, we have $x^TOx = {||Zx||^2}/{n}$ so that $O$ has the
same rank as $Z$, equal to $K$. Hence $O$ is invertible and positive
definite, thus the matrix $O^{1/2}$ (resp. its inverse) is well defined. 

\begin{proposition}\label{prop:FormEigen} Let $\mu \neq 0$ and $M_0=O^{1/2}BO^{1/2}$. The following statements are equivalent: 
 \begin{enumerate}
  \item $u=Zx$ is an eigenvector of ${A}$ associated to $\alpha_n \mu$.
  \item $O^{1/2}x$ is an eigenvector of $M_0$ associated to $\mu$;
 \end{enumerate}
In particular, the non-zero eigenvalues of $A$ are of the same order as  $\alpha_n$.
\end{proposition}

\subsection{Observed adjacency matrix}

In practice, we observe the adjacency matrix $\hat A$, which is as a noisy version of $A$. Our hope is that the $K$ leading eigenvectors of $\hat A$ are not too far from the 
$K$ leading eigenvectors of $A$, so that in view of Proposition~\ref{prop:Insight}, the solution in $Z'$ the following optimization problem provides a good estimate of $Z$:
$$
\min_{Z'\in \cZ,X'\in \R^{K\times K}} ||\hat U- Z'X'||_F,
$$
where $\hat U$ is the matrix  of $K$ normalized eigenvectors of $\hat A$ associated to the $K$ eigenvalues with largest absolute values.

This hope is supported by the following result on the perturbation of the leading eigenvectors of the adjacency matrix of any random graph. In practice, the 
number of communities $K$ is unknown and 
this result also 
provides an adaptive procedure to select the eigenvectors to use in the spectral embedding. Lemma~\ref{lem:FroebeniusAdjEmpOptim} relies on a combination of classical (algebraic) spectrum perturbation results with some (matrix) concentration inequalities. More details are provided in \ref{proof:Estimation}, where a proof for the following statement can be found.

We denote by $\lambda_{\min}(A)$ the smallest absolute value of a non-zero eigenvalue of $A$.

\begin{lemma}\label{lem:FroebeniusAdjEmpOptim} Let  $\delta\in ]0,1[$ and $\eta \in ]0,1/2[$. Let $\hat{U}$ be a matrix formed by orthogonal eigenvectors of $\hat{A}$ with an associated eigenvalue $\lambda$ that satisfy \[|\lambda| \geq \sqrt{2\left(1+\eta\right)\dmaxE \log(4n/\delta)}.\]
Let $\hat{K}$ be the number of such eigenvectors.
Let $U$ be matrix of $\hat{K}$ largest eigenvectors of $A$. If
\[ \dmax \geq \frac{4(2\eta+3)(2+\eta)}{3\eta^2}\log\left(\frac{4n}{\delta}\right)\ \ \ \text{and} \ \ \ \frac{\lambda_{\min}({A})^2}{\dmax}> \sqrt{2\left(1+\frac{\eta}{2+\eta}\right)}(1+\sqrt{1+\eta})\log \left(\frac{4n}{\delta}\right) ,\]
then with probability larger than $1-\delta$, $\hat{K}=\mathrm{rank}(A)$ and there exists a matrix $\hat{P}\in \cO_n(\R)$ such that  
\[\left|\left|\hat{U} - {U}\hat{P} \right|\right|_F^2 \leq {16K}\left(1+\frac{\eta}{\eta + 2}\right)
\left(\frac{\dmax}{\lambda_{\min}(A)^2}\right)\log\left(\frac{4n}{\delta}\right).\]
\end{lemma} 

Under SBMO, we have  $\lambda_{\min}({A})=\Theta(\alpha_n)$ by Proposition~\ref{prop:FormEigen}. As $\dmax = \Theta(\alpha_n)$, we need that $\alpha_n/\log(n)\rightarrow \infty$ to use Lemma~\ref{lem:FroebeniusAdjEmpOptim} to prove that $\hat{U}$ is a good estimate of $U$. We give in the next section sufficient conditions on the parameter $\alpha_n$ to obtain asymptotically exact recovery of the communities.

\section{The SAAC algorithm \label{sec:Algorithms}}

The spectral structure of the adjacency matrix suggests that $\hat{Z}$ defined below is a good  estimate of  the membership matrix $Z$ in the SBMO:
\begin{equation}
 \left(\cP\right) : \ \ (\hat{Z},\hat{X}) \in   \underset{{Z'\in \cZ, X' \in \R^{K\times K}}}{\text{argmin}} \ ||\hat{U}- Z'X'||_{F}^2, 
\label{FirstOptim}
 \end{equation}
 where  $\hat{U}\in \R^{n\times K}$ is the matrix of the $K$ normalized leading eigenvectors of $\hat{A}$.
In practice, solving $(\cP)$ is very hard, and the algorithm introduced in Section~\ref{subsec:Implementation} solves a relaxation of $(\cP)$ in which $Z'$ is only constrained to have binary entries, that is amenable to alternate minimization. In Section~\ref{subsec:Consistency}, we prove that an adaptive version of the estimate $\hat{Z}$ given by~\eqref{FirstOptim} is consistent.

\subsection{Description of the algorithm\label{subsec:Implementation}}

The \emph{spectral algorithm with additive clustering} (SAAC) consists in first computing a matrix $\hat{U}\in \R^{n\times K}$ whose columns are normalized eigenvectors of $\hat{A}$ associated to the $K$ largest eigenvalues (in absolute value), and then computing the solution of the following optimization problem: 
\[
 \left(\cP\right)' : \ \  (\hat{Z},\hat{X}) \in \underset{\substack{Z'\in \{0,1\}^{n\times {K}} : \forall i, Z'_i \neq 0\\X' \in \R^{K\times K}}}{\text{argmin}} \ ||\hat{U}-Z'X'||_{F}^2. 
 \]
$(\cP)'$ is reminiscent of the (NP-hard) $k$-means problem, in which the same objective function is minimized under the additional constraint that $||Z_i||=1$ for all $i$. 
The name of the algorithm highlights the fact that, rather than finding a clustering of the rows of $\hat{U}$, the goal is to find $\hat{Z}$, containing pure nodes $\hat{i}_1,\dots,\hat{i}_k$, that reveals the underlying additive structure of $\hat{U}$: for all $i$, $\hat{U}_i$ is not too far from $\sum_{k} \hat{U}_{\hat{i}_k} \ind_{(\hat{Z}_{i,k}=1)}$, in view of \eqref{AdditiveStructure}.

In practice, just like $k$-means, we propose to solve $(\cP)'$ by an alternate minimization over $Z'$ and $X'$. 
The proposed implementation  of the adaptive version of the algorithm, inspired by Theorem~\ref{thm:AnalysisASCO}, is presented as Algorithm~\ref{AlgoBox:ASCA}. 
An upper bound $m$ on the maximum overlap $O_{\max}=\max\{|| z ||, z\in \cT\}$ is provided to limit the combinatorial complexity of the algorithm. If $K$ if known, the selection phase can be removed, and one use directly the matrix $\hat{U} \in \R^{n\times K}$ of $K$ leading eigenvectors.  While heuristics do exist for selecting the number of clusters in spectral clustering (e.g. \cite{VLux08Tuto,Zelnik}), this thresholding procedure is supported by theory for networks drawn under SBMO. It is reminiscent of the USVT algorithm of  \cite{Chatterjee14USVT}, that can be used to estimate the expected adjacency matrix in a SBM.

\begin{algorithm}[ht]
  \caption{Adaptive SAAC\label{AlgoBox:ASCA}}
  \begin{algorithmic}[1]
    \REQUIRE Parameters $\epsilon$, $r$, $\eta>0$. Upper bound $m$ on the maximum overlap $O_{\max}$.
    \REQUIRE $\hat{A}$, the adjacency matrix of the observed graph.
    \STATE $\sharp$ Selection of the eigenvectors
    \STATE Form $\hat{U}$ a matrix whose columns are $\hat{K}$ eigenvectors of $\hat{A}$ associated to eigenvalues $\lambda$ satisfying  
   \[ |\lambda| > \sqrt{2(1+\eta)\hat{d}_{\max}\log(4n^{1+r})}\]  
    
    \vspace{-0.3cm}
    \STATE $\sharp$ Initialization
    \STATE $\hat Z=0 \in \R^{n\times \hat{K}}$
    \STATE  $\hat X\in \R^{\hat{K}\times \hat{K}}$  initialized with $k$-means++  applied to $\hat U$, the first centroid being chosen at random among nodes with  degree smaller than the median degree
    \STATE $\emph{Loss}=+\infty$
    \STATE $\sharp$ Alternating minimization   
    \WHILE{($\emph{Loss} - ||\hat{U}-\hat Z \hat X||^2_{F} > \epsilon$)} 
   \STATE $\emph{Loss}=||\hat{U}-\hat Z \hat X||^2_{F}$
    \STATE Update membership vectors: 
    $\forall i, \  \hat Z_{i,\cdot} = \underset{{z\in \{0,1\}^{1\times \hat{K}}: 1 \leq ||z||_1 \leq m}}  {\arg\min} \ ||\hat{U}_{i,\cdot} - z \hat X||.$
   \STATE Update centroids:
    $\hat X =(\hat Z^T\hat Z)^{-1}\hat Z^T\hat{U}.$
    \ENDWHILE
  \end{algorithmic}
\end{algorithm}
 
 Alternate minimization is guaranteed to converge, in a finite number of steps, towards a local minimum of  $||\hat Z\hat X - \hat{U} ||_{F}^2$. However, the convergence is very sensitive to initialization. We use a $k$-means$++$ initialization (see \cite{kmeansplusplus}), which is a randomized procedure that picks as initial centroids rows from $\hat{U}$ that should be far from each other. For the first centroid, we choose at random a row in $\hat{U}$ corresponding to a node whose degree is smaller than the median degree in the network. We do so because in the SBMO model, pure nodes tend to have smaller degrees and we expect the algorithm to work well if the initial centroids are chosen not too far from rows in $\hat{U}$ corresponding to pure nodes. 

Given $\hat Z$, as long as the matrix $\hat Z^T\hat Z$ is invertible, there is a closed form  solution to the minimization of  $||\hat Z \hat X -\hat U||_F$ in $\hat X$, which is $\hat X= (\hat Z^T\hat Z)^{-1}\hat Z^T\hat U = \hat{Z}^+ \hat U$, where $\hat{Z}^+$ is the pseudo-inverse of $Z$. The fact that $\hat Z^T\hat Z$ is not invertible implies in particular that $\hat Z$ does not contain a pure node for each community.
If this happens, we  re-initialize the centroids, using again the $k$-means$++$ procedure.

Computing the spectral embedding can be done efficiently even for large graphs, as one can use variants of the power method for the sparse matrix $A$ (see, e.g. \cite{MatrixComputations}). Then, the alternating minimization part of the algorithm usually converges in a few step. However in each step updating the membership vectors can be quite expensive, as it requires for each node to enumerate all possible membership vectors in $\{0,1\}^K$ with support smaller than $m$. Thus in practice, when the number of communities is large, the parameter $m$ limiting the size of the overlap should not be set too large for the algorithm to be computationally efficient.

\subsection{Consistency of an adaptive estimator \label{subsec:Consistency}}

We give in Theorem~\ref{thm:AnalysisASCO} theoretical properties for a slight variant of the estimate $\hat{Z}$ in \eqref{FirstOptim}, that is solution of the optimization problem $(\cP_\epsilon)$ defined therein, that features the set of membership matrices for which the proportion of pure nodes in each community is larger than $\epsilon$:  
\[\cZ_{\epsilon} =\left\{ Z'\in \{0,1\}^{n\times K},\:\forall k\in\{1,\cdots,K\}\:, \frac{|\{ i : Z'_{i}=\ind_{\{k\}}\}|}{n} > \epsilon \right\}.
\]
Recall the notation introduced in~\eqref{def:BetaL} and \eqref{def:Lz}. We assume that $\epsilon$ is smaller than the smallest proportion of pure nodes (in the limit), given by $\min_{k} \beta_{\ind_{\{k\}}}$, and let $L_{\max}=\max_{z}L_z$. 

The estimator analyzed is adaptive, for it relies on an estimate $\hat{K}$ of the number of communities, and on $\hat{\cZ}_\epsilon = \cZ_{\epsilon}(\hat{K})$.  We establish its consistency for any fixed matrices $B$ and $Z$ satisfying (SBMO1) and (SBMO2). It is to be noted that while the consistency result for the OCCAM algorithm \cite{Levina14Overlap} applies to moderately dense graphs ($\alpha_n$ has to be of order $n^\alpha$ for some $\alpha>0$), our result handle relatively sparse graphs, in which $\alpha_n$ is of order $(\log(n))^{1+c}$ for some $c>0$. Our result involves constants defined below, that are related to the overlap matrix $O$ and to the  matrix $O^{1/2}BO^{1/2}$ introduced in Proposition \ref{prop:FormEigen}.

\begin{definition}\label{def:KeyQuantities} The core matrix is the $K\times K$ symmetric matrix $M_0 := O^{1/2} B O^{1/2}$. We let 

\vspace{-0.3cm}

\begin{eqnarray*}
\mu_0 &:=& \min \{ |\lambda| : \lambda\neq 0 \ \text{is an eigenvalue of} \ M_0\},\\
d_0   &:=& \min_{\substack{z \in \{-1,0,1,2\}^{1\times K}\\ z\neq 0}}
\left|\left|z O^{-1/2}\right|\right|>0. 
\end{eqnarray*}
\end{definition}

Note that $d_0$ is positive as seen by the following argument:
if $d_0=0$, then there would exist a linear combination of the rows of $O^{-1/2}$ which is zero; this is
impossible because the matrix $O^{-1/2}$ is invertible. 

\begin{theorem}\label{thm:AnalysisASCO} Let $\eta \in ]0,1/2[$ and
  $r>0$. Let $\hat{U}$ be a matrix whose columns are orthogonal
  eigenvectors of $\hat{A}$ associated to an eigenvalue ${\hat \lambda}$ satisfying \[|{\hat \lambda}| \geq \sqrt{2\left(1+\eta\right)\dmaxE \log(4n^{1+r})}.\]
Let $\hat{K}$ be the number of such eigenvectors.
Let  
\[
 (\cP_\epsilon): \ \ \ \ (\hat{Z},\hat{X}) \in \underset{Z'\in \cZ_{\epsilon}, X' \in \R^{\hat{K}\times \hat{K}}}{\text{argmin}} \ ||Z'X' - \hat{U} ||_{F}^2.
\]
Assume that $\frac{\alpha_n}{\log n}\to \infty$ and $ \min_k \beta_{\ind_{\{k\}}} > \epsilon$. There exists some constant $c_1>0$ such that, if 
\[\alpha_n \geq \max\left[\frac{4(2\eta+3)(2+\eta)}{3\eta^2L_{\max}} ; \sqrt{2\left(1+\frac{\eta}{2+\eta}\right)}\frac{1+\sqrt{1+\eta}}{\mu_0^2}\right]\log \left(4n^{1+r}\right)\]
then, for $n$ large enough, with probability larger than $1-n^{-r}$, $\hat{K}=K$ and 
\[
\frac{\MisC(\hat{Z},Z)}{n} \leq c_1\frac{K^3L_{\max}}{d_0^2\mu_0^2}
\frac{\log(4n^{1+r})}{\alpha_n}.
\]
\end{theorem}

In particular, assuming that  ${\alpha_n}/{\log(n)}{\rightarrow} \infty$ when $n$ goes to infinity, it can be shown (using the Borel-Cantelli Lemma) that
the estimation procedure described in Theorem~\ref{thm:AnalysisASCO} with a  parameter $r\geq 2$ is consistent, in the sense that it satisfies  
\[
\frac{\MisC(\hat{Z},Z)}{n} \underset{n\rightarrow \infty}{\overset{a.s.}{\longrightarrow}} 0.
\]

\paragraph{Theoretical guarantees for other estimates} Theorem~\ref{thm:AnalysisASCO} leads to an upper bound on the estimation error of $\hat{Z}$ a solution to $(\cP_\epsilon)$. In some cases, it is also possible to prove directly that the solution of $(\cP)'$ leads to a consistent estimate of $Z$. This is the case for instance in an identifiable SBMO with two overlapping communities or with three communities with pairwise overlaps. 

If $K$ is known, tighter results can be obtained for non-adaptive procedures in which  $\hat{U}\in\R^{n\times \hat{K}}$ is replaced by $\hat{U}\in\R^{n\times K}$. These results are stated in~\ref{sec:NonAdaptive}, where two non-adaptive estimation procedures are shown to be consistent under the (looser) condition $\alpha \geq c_0\log(n)$ for some constant $c_0$ stated therein.

\subsection{Proof of Theorem~\ref{thm:AnalysisASCO}\label{sec:Proofs}}

Let $U \in \R^{n\times K}$ be a matrix whose columns are $K$ independent normalized eigenvectors of $A$ associated to the non-zero eigenvalues. The proof strongly relies on the following decomposition of $U$, that is a consequence of Proposition~\ref{prop:FormEigen}. 

\begin{lemma}\label{prop:Decomposition} There exists a matrix $V\in \cO_K(\R)$ of eigenvectors of $M_0=O^{1/2}BO^{1/2}$ such that 
$U= ZX$ with $X=n^{-1/2}O^{-1/2}V$.
\end{lemma}

We state below a crucial result characterizing the sensitivity to noise of the decomposition $U=ZX$ of Lemma~\ref{prop:Decomposition}, in terms of the quantity $d_0$ introduced in Definition~\ref{def:KeyQuantities}. The proof of this key result is given in~\ref{sec:proofSensitivity}: it builds on fact that $d_0$ provides a lower bound on the norm of some particular linear combinations of the rows of $X$: indeed, one has 
\[\forall \ z \in \{-1,0,1,2\}^{1\times K} \backslash \{0\}, \ \ \ ||zX|| \geq {d_0}/{\sqrt{n}}.\]

\begin{lemma}\label{prop:Robustness} (Robustness to noise)  Let $Z'\in \R^{n\times K}$, $X'\in \R^{K\times K}$ and $\cN \subset \{1,\dots,n\}$. Assume that 
\begin{enumerate}
 \item $\forall i \in \cN$, $||Z_i'X' - U_i|| \leq \frac{d_0}{4K\sqrt{n}}$   
  \item there exists $(i_1,\dots,i_K),(j_1,\dots,j_K)\in(\cN^K)$: $\forall k \in [1,K], \ Z_{i_k}=Z'_{j_k}=\ind_{\{k\}}$
\end{enumerate}
Then there exists a permutation matrix $P_\sigma$ such that for all $i \in \cN$, $Z_i = (Z'P_\sigma)_i$.
\end{lemma}

Let $\hat{U}$ the matrix defined in Theorem~\ref{thm:AnalysisASCO}. We first note that Lemma~\ref{lem:FroebeniusAdjEmpOptim} can be rephrased in terms of the degree parameter $\alpha_n$. Indeed, from Proposition~\ref{prop:FormEigen}, $\lambda_{\min}(A)=\alpha_n \mu_0$, with $\mu_0$ in Definition~\ref{def:KeyQuantities} and $\dmax=\alpha_n L_{\max}$, with  
\[L_{\max} = \max_{i=1\dots n} \left(\frac{1}{n}Z_iBZ^T\ind_{n,1}\right).\]
From Lemma~\ref{lem:FroebeniusAdjEmpOptim}, letting 
 \[ C_0(\eta)=\max\left[\frac{4(2\eta+3)(2+\eta)}{3\eta^2L_{\max}} ; \sqrt{2\left(1+\frac{\eta}{2+\eta}\right)}\frac{1+\sqrt{1+\eta}}{\mu_0^2}\right],\]
if $\alpha_n \geq C_0(\eta)\log \left({4n^{1+r}}\right)$ then with probability larger than $1-n^{-r}$, $\hat{K} = K$ and there exists a rotation $\hat{P} \in \cO_K(\R)$ such that  
\begin{equation}||\hat{U} - U \hat{P}||_F^2 \leq 16K\left(1 + \frac{\eta}{\eta +2}\right)\frac{L_{\max}}{\mu_0^2}\left(\frac{\log(4n^{1+r})}{\alpha_n}\right)\label{FinalUB}.\end{equation}
In the sequel, we assume that $\hat{K}=K$ and that this inequality holds with a rotation $\hat{P}$. 

The estimate $\hat{Z}$,$\hat{X}$ is then defined by 
\[(\hat{Z},\hat{X}) \in  \underset{{Z'\in \cZ_{\epsilon}(K),   X'\in \R^{K\times K}}}{\text{argmin}} \ ||Z'X' - \hat{U} ||_{F}^2.
\]
Introducing $\hat{X}_1:=\hat{X}\hat{P}^{-1}$, we first show that $\hat{Z}\hat{X}_1$ is a good estimate of $U$ provided that $\hat{U}$ is:
\begin{equation}||\hat{Z}\hat{X}_1 - U||_F \leq 2 ||U\hat{P} - \hat{U}||_F.\label{ZCEstimate}\end{equation}
This inequality can be obtained in the following way. Let $X,Z$ be defined in Lemma~\ref{prop:Decomposition}. As $Z\in\cZ_{\epsilon}$ (for $\epsilon < \min_k \beta_{\ind_{\{k\}}}$), by definition of $\hat{Z}$ and $\hat{X}$, 
\[||\hat{Z}\hat{X} - \hat{U} ||_{F}^2 \leq || ZX\hat{P} - \hat{U} ||_{F}^2 = || U\hat{P} - \hat{U} ||_{F}^2.\]
Then, one has 
\begin{eqnarray*}
||\hat{Z}\hat{X}\hat{P}^{-1} - U||_{F} & \leq & ||\hat{Z}\hat{X}\hat{P}^{-1} - \hat{U}\hat{P}^{-1} ||_{F} + ||\hat{U}\hat{P}^{-1} - U ||_{F}   = ||\hat{Z}\hat{X} - \hat{U}||_{F} + ||\hat{U} - U \hat{P}||_{F} \\
&\leq&  2||\hat{U} - U\hat{P}||_{F}. 
\end{eqnarray*}

We now introduce the set of nodes
\[\cN_n=\left\{ i : ||\hat{Z}_i\hat{X}_1 - U_i|| \leq \frac{d_0}{4K\sqrt{n}}\right\} \]
and show that assumption 1. and 2. in Lemma~\ref{prop:Robustness} are satisfied for this set and the pair  $(\hat{Z},\hat{X}_1)$, if
\begin{equation}
\frac{64K^2}{d_0^2}||\hat{U} - {U}\hat{P}||_F^2 \leq \epsilon.
 \label{AssumptionCdt}
\end{equation}
Assumption 1. is satisfied by definition of $\cN_n$. We now show that, as required by assumption 2., $\cN_n$ contains one pure node in each community relatively to $Z$ and $\hat{Z}$.

First, using notably \eqref{ZCEstimate}, the cardinality of $\cN_n^c$ is upper bounded as
\[
 \frac{|\cN_n^c|}{n}  =  \frac{\sum_{i\in\cN_n^c} \!\!1}{n} \leq \frac{16K^2}{d_0^2}\sum_{i=1}^n ||\hat{Z}_i\hat{X}_1 - U_i||^2  = \frac{16K^2}{d_0^2}||\hat{Z}\hat{X}_1 - U||_F^2 
 \leq  \frac{64K^2}{d_0^2}||\hat{U} - {U}\hat{P}||_F^2. \nonumber
\]
Thus, if \eqref{AssumptionCdt} holds, ${|\cN_n^c|}\leq \epsilon n$. As $\hat{Z} \in \cZ_{\epsilon}(K)$, for all $k\leq K$ the cardinality of the set of nodes $i$ such that $\hat{Z_i}=\ind_{\{k\}}$ is strictly larger than $\epsilon n$, hence this set cannot be included in $\cN_n^c$. Thus, for all $k$, there exists $j_k\in\cN_n$ such that $\hat{Z}_{j_k}=\ind_{\{k\}}$. As $\epsilon$ is smaller than $\min_{k}\beta_{\ind_{\{k\}}}$, the minimal proportion of pure nodes in a community, by a similar argument the set of nodes $i$ such that $Z_i = \ind_{\{k\}}$ cannot be included in $\cN_n^c$ either. Thus for all $k$, there exists $i_k\in\cN_n$ such that $Z_{i_k}=\ind_{\{k\}}$. 

Hence Lemma~\ref{prop:Robustness} can be applied and there exists $\sigma\in \mathfrak{S}_K$ such that 
$\forall i \in \cN_n, \ \ \hat{Z}_{i,\sigma(k)}=Z_{i,k}$: up to a permutation of the community labels, all the communities of nodes in $\cN_n$ are recovered. Using \eqref{FinalUB}, this implies that whenever $\alpha_n \geq C_0(\eta)\log \left({4n^{1+r}}\right)$,  with probability larger than $1 - n^{-r}$,
\[\frac{\MisC(\hat{Z},Z)}{n} \leq \frac{|\cN_n^c|}{n} \leq  \frac{64K^2}{d_0^2}||\hat{U} - {U}\hat{P}||_F^2\leq \frac{1024K^3L_{\max}}{d_0^2\mu_0^2}\!\left(\!1\!+\!\frac{\eta}{\eta\! +\! 2}\!\right)\!
\frac{\log(4n^{1+r})}{\alpha_n},\]
provided that the final upper bound is smaller that $\epsilon$ (which implies that the condition \eqref{AssumptionCdt} is satisfies), which is the case for $n$ large enough.

\section{Experimental results}\label{sec:Experiments}

We mostly use the estimation error to evaluate the quality of an estimate $\hat{Z}$ of some membership matrix $Z$, that we recall is defined by 
\[\text{Error}(\hat{Z},Z) = \frac{1}{nK} \min_{\sigma \in \mathfrak{S}_K} ||\hat{Z}P_\sigma - Z ||_F^2.\]
This error can be split into two kinds of errors: entries that are ones in $\hat{Z}P_{\sigma^*}$ (where $\sigma^*$ realizes the minimum above) but zeros in $Z$, called false positive, and entries that are zeros in $\hat{Z}P_{\sigma^*}$ but ones in $Z$, called false negative. We define the false positive and false negative rates as 
\[\text{FP}(\hat{Z},Z) = \frac{ |(i,k) : \hat{Z}_{i,\sigma^*(k)} = 1 \ \text{and} \ Z_{i,k}=0|}{ |(i,k) : \ Z_{i,k}=1|},  \ \ \text{FN}(\hat{Z},Z) = \frac{ |(i,k) : \hat{Z}_{i,\sigma^*(k)} = 0 \ \text{and} \ Z_{i,k}=1|}{|(i,k) : \ Z_{i,k}=0|}.\]

An extension of the \emph{normalized variation of information} (NVI) introduced by \cite{Lancichinetti09Eval} is also used as a measure of performance in several papers. This indicator compares the distribution of two random vectors $\bm{X}=(X_1,\dots,X_K)$ and $\bm{Y}=(Y_1,\dots,Y_K)$ in $\{0,1\}^K$ associated to $\hat{Z}$ and $Z$ respectively, such that the joint distribution of any two marginal is given by 
\[\bP(X_k = x_k, Y_l = y_l) = \frac{|i : \hat{Z}_{i,k}=x_k \ \text{and} \  Z_{i,l}=y_l |}{n}.\]

\vspace{-0.3cm}

\hspace{-0.5cm}The NVI is defined by
\[\text{NVI}(\hat{Z},Z) = 1 - \min_{\sigma \in \mathfrak{S}_K} \frac{1}{2K}\sum_{k=1}^K \left[\frac{H(X_k | Y_{\sigma(k)})}{H(X_k)} + \frac{H(Y_{\sigma(k)}|X_k)}{H(Y_{\sigma(k)})}\right],\]
where $H(V)$ and $H(V|W)$ denote respectively the entropy of a random variable $V$ and the conditional entropy of $V$ given $W$ (see e.g. \cite{Cover:Thomas} for definition). Unlike the other performance measures that we consider, the NVI should be maximized.

Our analysis shows that for a graph drawn under the SBMO the error of SAAC goes to zero almost surely when the number of nodes $n$ grows large, and the degrees are large enough, more precisely (slightly more than) logarithmic in $n$. We illustrate this fact on simulated data, and compare SAAC to other (spectral) algorithms on simulated data and on two kinds of real-world graphs with overlapping communities : ego networks and co-authorship networks.

\subsection{Simulated data}

We compare SAAC to (normalized) spectral clustering using the adjacency matrix, referred to as SC and to the spectral algorithm proposed by \cite{Levina14Overlap} to fit the random graph model called OCCAM. We refer to this algorithm as the OCCAM spectral method. 

\begin{figure}[ht]
\centering
\includegraphics[width=0.52\textwidth]{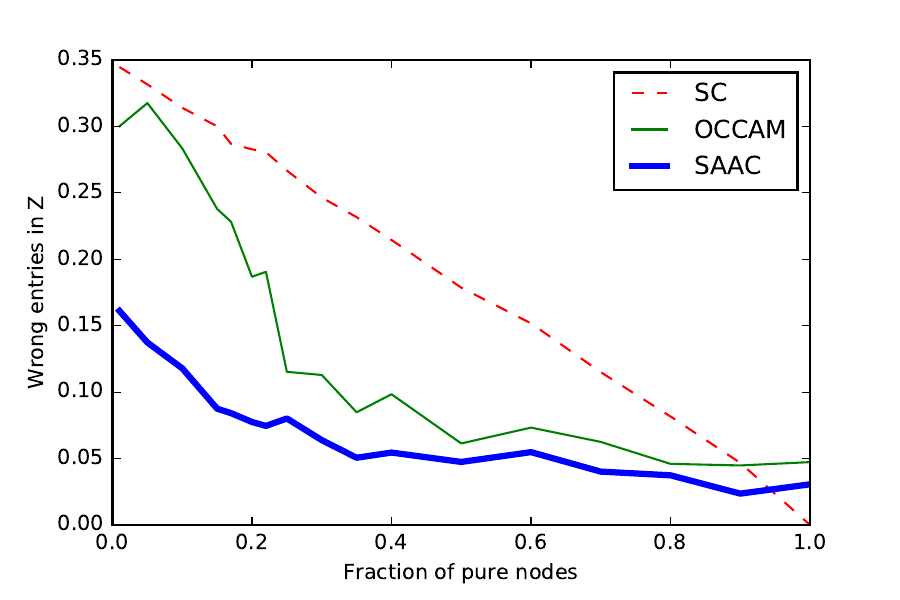}
\hspace{-1cm}
\includegraphics[width=0.52\textwidth]{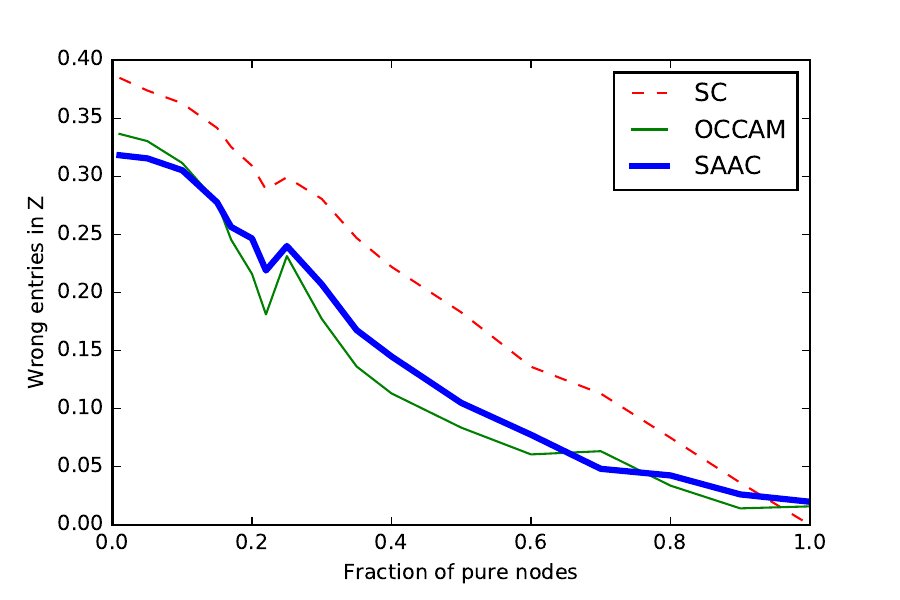}
\caption{\label{Simuoverlap} Comparison of SC, SAAC and the OCCAM spectral algorithm under instances of  SBMO (left) and  OCCAM (right) random graph models.}
\end{figure}

First, we generate networks from SBMO models with $n=500$ nodes, $K=5$ communities, $\alpha_n=\log^{1.5}(n)$, $B=\text{Diag}([5,4,3,3,3])$ and $Z$  drawn at random in such a way that each community has a fraction of pure nodes equal to $p/K$ for some parameter $p$ and the size of the maximum overlap $O_{\max}$ is smaller than $3$. 
The left part of Figure~\ref{Simuoverlap} shows the error of each method as a function of $p$, averaged over 100 networks. SAAC significantly outperforms OCCAM, especially when there is a 
large overlap between communities. As expected, both methods outperform SC, which is designed to handle non-overlapping communities, except when the amount of overlap gets really small. 

To have a more fair comparison with the OCCAM spectral algorithm, we then draw networks under a modified version of the model used before, in which the rows of $Z$ are normalized, so that for all $i$, one has $||Z_i||=1$: this random graph model is a particular instance of the OCCAM. Results are displayed on the right part of Figure~\ref{Simuoverlap}. The OCCAM spectral algorithm, designed to fit this model, performs most of the time slightly better than the other methods, but the gap between OCCAM and SAAC is very narrow. 

\vspace{-0.3cm} 
 
\subsection{Real networks}

\cite{Levina14Overlap} compare the performance of the  OCCAM spectral algorithm to that of other algorithms on both simulated data and real data, namely ego networks \cite{EgoSNAP}. Nodes in an ego network are the set of friends of a given central node in a social network, and edges indicate friendship relationships between these nodes. 
The ground-truth communities, corresponding to circles of friends and asked to the users themselves,  are available \cite{snap}.
We first apply SAAC on networks from this dataset, that naturally contain overlap. To do so, we use the pre-processing of the networks described in \cite{Levina14Overlap}, that especially keeps communities if they have at least a fraction of pure nodes equal to 10\% of the network. Additionally, because the focus is on overlapping communities, we keep only networks for which the fraction of nodes that belong to more than one community is larger than 1\%. This leads us to keep only 6 (out of 10) Facebook networks (labeled 0, 414, 686, 698, 1912 and 3437 in the 
dataset), 26 (out of 133) Google Plus networks from the original dataset (labeled 8, 11, 18, 28, 31, 37, 39, 43, 44, 52, 66, 67, 69, 75, 76, 89, 91, 94, 95, 100, 106, 108, 112, 116, 129, and 130) and 96 networks (out of 973) from the Twitter dataset.    

\begin{table}[p]
\centering
\begin{tabular}{|c|c|c|c|}
\hline
  $n$   & $K$    &  c     & $O_{\max}$ \\
 \hline
        190   &  3.17  & 1.09   & 2.17       \\
           (173) & (1.07) & (0.06) & (0.37)     \\
\hline
\end{tabular}
\caption{\label{TableFB0} Characteristics of  the considered ego-networks from Facebook.}
\end{table}

\begin{table}[p]
\centering
\begin{tabular}{|c|c|c|c|c|c|c|}
\hline
  &  FP     & FN      & Error   & NVI    & $\hat{c}$ & $\hat{O}_{\max}$\\
\hline
SC              & 0.108 & 0.127 & 0.109 & 0.578    & 1 & 1 \\
            & (0.107) & (0.103) & (0.080) &  (0.257)&   & \\
\hline
OCCAM          & 0.171 & 0.102 & 0.119 & 0.574  & 1.11 & 2.5\\
           & (0.183) & (0.078) & (0.103) & (0.282) & (0.082) & (0.5) \\
\hline
SAAC            & 0.116 & 0.102 & 0.100 & 0.550  & 1.06 & 2 \\
       & (0.066) & (0.065) & (0.052) & (0.224)  & (0.055) & (0.577) \\
\hline
\end{tabular}
\caption{\label{TableFB} Spectral algorithms recovering overlapping friend circles in ego-networks from Facebook.}

\end{table}

\begin{table}[p]
\centering
\begin{tabular}{|c|c|c|c|}
\hline
  $n$   & $K$    &  c     & $O_{\max}$ \\
 \hline
        467   &  2.65  & 1.10  & 2.31       \\
           (431) & (0.95) & (0.10) & (0.67)     \\
\hline
\end{tabular}
\caption{\label{TableGPlus0} Characteristics of  the considered ego-networks from Google Plus.}
\end{table}

\begin{table}[p]
\centering
\begin{tabular}{|c|c|c|c|c|c|c|}
\hline
  &  FP     & FN      & Error   & NVI     & $\hat{c}$ & $\hat{O}_{\max}$\\
\hline
SC & 0.151 &  0.194 & 0.163 & 0.453 & 1 & 1\\
   & (0.118) &  (0.122) & (0.103) & (0.214) &  & \\
\hline 
OCCAM & 0.252 & 0.113 & 0.154 &  0.482 & 1.223 & 2.538  \\
& (0.219) & (0.106) &  (0.109) &  (0.221) & (0.206) &  (0.887)\\
\hline
SAAC &  0.257 & 0.127 & 0.174 & 0.448 & 1.196 & 2.308 \\
& (0.188) & (0.110) & (0.116) & (0.220) &  (0.176) & (0.666)\\
\hline
\end{tabular}
\caption{\label{TableGPlus} Spectral algorithms recovering overlapping friend circles in ego-networks from Google Plus.}

\end{table}

\begin{table}[p]
\centering
\begin{tabular}{|c|c|c|c|}
\hline
  $n$   & $K$    &  c     & $O_{\max}$ \\
 \hline
        68   &  3.34  & 1.09  &   2.14     \\
        (35)    & (0.96)  & (0.09)  & (0.37)    \\
\hline
\end{tabular}
\caption{\label{TableTwitter0} Characteristics of  the considered ego-networks from Twitter.}
\end{table}

\begin{table}[p]
\centering
\begin{tabular}{|c|c|c|c|c|c|c|}
\hline
  &  FP     & FN      & Error   & NVI     & $\hat{c}$ & $\hat{O}_{\max}$\\
\hline
SC & 0.255 &  0.181 &  0.193 &  0.351 & 1 & 1\\
   & (0.143) &  (0.107) &  (0.093) &  (0.199) &  & \\
\hline 
OCCAM & 0.492 & 0.124 & 0.225 & 0.330 & 1.383 &  2.792 \\
&  (0.302) & (0.099) & (0.104) & (0.203) & (0.261) &  (0.720) \\
\hline
SAAC &  0.446 &  0.131 & 0.222 & 0.349 & 1.303 & 2.104 \\
&  (0.253) & (0.090) & (0.104) & (0.205) & (0.209) & (0.395)\\
\hline
\end{tabular}
\caption{\label{TableTwitter} Spectral algorithms recovering overlapping friend circles in ego-networks from Twitter.}

\end{table}

Tables~\ref{TableFB0} and \ref{TableFB} present respectively  the characteristics of the Facebook networks used and the performance of SC, SAAC and OCCAM, averaged over the 6 networks used (with the standard deviation added). For each algorithm, the estimation error is displayed but also the fraction of false positive (FP) and false negative (FN) entries in $\hat{Z}$, and the extended normalized variation of information (NVI). The parameter $c$ corresponds to the average number of communities per node, $c=\sum_{i,k} Z_{i,k}/n$ and $O_{\max}$ is the maximum size of an overlap. OCCAM and SAAC have comparable performance, but there is no significant improvement over spectral clustering. This can be explained by the fact that the amount of overlap ($c$) is very small in this dataset. The same tendency was observed on the Google Plus networks, for which the results of our experiments are displayed in Tables~\ref{TableGPlus0} and \ref{TableGPlus} as well as on the Twitter networks, for which the results can be found in Tables~\ref{TableTwitter0} and \ref{TableTwitter}. We also report the  average number of communities by node $\hat{c}$ and maximum size of overlap $\hat{O}_{\max}$ found by each algorithm (averaged over all networks). One can see that OCCAM is slightly more prone to overestimate the amount of overlap compared to SAAC.

We then try SAAC on co-authorship networks built from DBLP in the following way. Nodes correspond to authors and we fix as ground-truth communities some conferences (or group of conferences): an author belongs to some community  if she/he has published at least one paper in the corresponding conference(s). We then build the network of authors by putting an edge between authors if they have published a paper together in one of the considered conferences. We present results for some conferences with machine learning in their scopes : ICML, NIPS, and two theory-oriented conferences that we group together, ALT and COLT. We compare the three spectral algorithms in terms of estimation error and false positive / false negative rates. Results are presented in 
Table~\ref{fig:ML}, in which the estimated amount of overlap $\hat{c}=\sum_{i,k}\hat{Z}_{i,k}/n$ is also reported. In this case, SAAC and OCCAM significantly outperform SC, although the error is relatively high. The amount of overlap is under-estimated by both algorithms, but SAAC appears to recover slightly more overlapping nodes. The difficulty of recovering communities in that case may come from the fact that the networks constructed are very sparse.

\begin{table}[ht]

\begin{minipage}{0.5\textwidth}
\[\cC_1=\{\emph{ICML}\}, \ \cC_2=\{\emph{ALT}, \emph{COLT}\}.\]

\vspace{-0.5cm}

\[n=4374, \ K =2, \ d_{\text{mean}}=3.8,\ c = 1.09\]
\centering
\begin{tabular}{|c|c|c|c|c|}
\hline 
             & $\hat{c}$ &  FP        &    FN       &  Error \\
\hline
SC        & 1.        &     0.39    &  0.55     &  0.46\\		
\hline
OCCAM     & 1.00    &   0.2    &  0.34     &  0.26\\		
\hline
SAAC        & 1.03   &    0.21    &  0.31     &  0.25\\	
\hline
\end{tabular}
\end{minipage}
\hspace{-0.2cm}
\begin{minipage}{0.5\textwidth}

\vspace{-0.4cm}

\[\cC_1=\{\emph{NIPS}\}, \ \cC_2=\{\emph{ICML}\},\cC_3=\{\emph{ALT}, \emph{COLT}\}\]

\vspace{-0.8cm}

\[n = 9272, \ K = 3, \ d_{\text{mean}} = 4.5,\ c=1.22\]
\centering
\begin{tabular}{|c|c|c|c|c|}
\hline 
        & $\hat{c}$ &  FP     & FN       & Error \\
\hline
SC &  1.        &   0.38 &  0.39  & 0.39\\
\hline
OCCAM  & 1.02    &   0.25 &  0.28 & 0.27\\
\hline
SAAC   & 1.04    &   0.26 &  0.28  & 0.27\\		
\hline
\end{tabular}
\end{minipage}

\vspace{0.1cm}

\caption{\label{fig:ML} Spectral algorithms recovering overlapping machine learning conferences}

\end{table}

\section{The sparse case\label{sec:ExpBonus}}

In a stochastic blockmodel the sparse regime, in which each node has a constant degree, has been  extensively studied over the past few years. A threshold under which it is not possible to do better than random guessing the communities has been identified. Moreover it is known that spectral clustering cannot be used close to this threshold and that more sophisticated methods are needed to detect communities. In this section we investigate on a simple example what happens in the SBMO. 

\newpage

Consider the following simple SBMO with two communities and a diagonal connectivity matrix such that if
$\mathbf{1}_{r}\in \R^{r\times 1}$ is a vector containing only ones, the expected adjacency matrix is 
\[A=\frac{\alpha_n}{n}ZBZ^T, \ \ \text{with} \ \  B= \left(\begin{array}{cc}
a&b\\
b&a
\end{array}\right) 
 \ \ \text{and} \ \ Z = \left(\begin{array}{cc}
\mathbf{1}_{sn}&0\\
\mathbf{1}_{(1-2s)n}&\mathbf{1}_{(1-2s)n}\\
0&\mathbf{1}_{sn}
\end{array}\right),\]
where $0<s<1/2$ is the fraction of pure nodes in each of the two communities: the smaller $s$, the larger the overlap, whereas $s=1/2$ corresponds to pure nodes only, i.e. a SBM without overlap. 
The matrix $A$ has rank $2$ with two non zero eigenvalues $\alpha_n (2-3s)(a+b)>\alpha_n s(a-b)$. The associated eigenvectors are respectively
\[
X=\left(\begin{array}{c}
\mathbf{1}_{sn}\\
\mathbf{2}_{(1-2s)n}\\
\mathbf{1}_{sn}
\end{array}
\right) \ \ \mbox{ and} \ \ \ Y=\left(\begin{array}{c}
\mathbf{-1}_{sn}\\
\mathbf{0}_{(1-2s)n}\\
\mathbf{1}_{sn}
\end{array}
\right).
\]
Each node $i$ of the network has a spectral embedding given by
$(X_i,Y_i)$, i.e. pure nodes in community one correspond to
$P_1=(1,-1)$, pure nodes in community two correspond to $P_2=(1,1)$
and mixed nodes correspond to $M=(2,0)$. As expected, we have
$M=P_1+P_2$ and if $\alpha_n$ is sufficiently large, i.e. $\alpha_n >>
\log n$, then Theorems \ref{thm:AnalysisASCO} (and
\ref{thm:AnalysisASCONonAdaptive} in the appendix) apply: the eigenvectors of the
empirical adjacency matrix $\hat{A}$ will be close to the eigenvectors
$X$ and $Y$ and as a consequence, the fraction of nodes that are
misclassified by SAAC will vanish as $n$ tends to infinity.

Let now consider the very sparse case where $\alpha_n=1$. In this case, the average degree in the graph is constant (i.e. not scaling with $n$) and our theoretical results are not valid. Indeed, we believe that there is a range of parameters where only partial recovery is possible.
Note that if one has only access to the eigenvector $X$ (associated to the largest eigenvalue $(2-3s)(a+b)$), then it is possible to distinguish pure nodes from mixed nodes but it is impossible to distinguish pure nodes of community one from pure nodes of community two. Observe that the second eigenvalue of $A$, $s(a-b)$ can be very small in which case this eigenvalue will be `hidden' in the noise of the model. 
In the sparse regime, it is known that high-degree nodes induce a lot
of noise on the spectrum of the adjacency matrix. 

\begin{figure}[ht]
\centering
\includegraphics[width=0.65\textwidth]{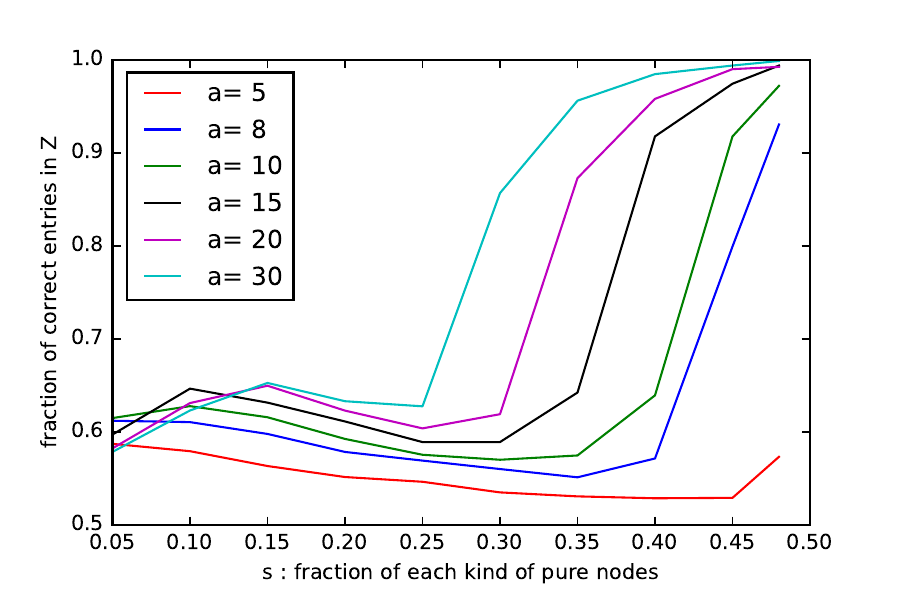}
\caption{\label{2B2overlap} Performance of SAAC as a function of the fraction $s$ of each type of pure nodes in a SBMO model with two-by-two overlap between $K=2$ 
communities}
\end{figure}

Figure~\ref{2B2overlap} illustrates how the performance of the SAAC algorithm deteriorates when the sparsity increases. 
 The fraction of correct entries
(1-$\Err(\hat{Z},Z)$) is displayed as a function of the parameter
$s$. The number of nodes is fixed to $n=1000$ and the curves in different colors correspond to different values of
$a$, the parameter $b$ being set to zero. For each value of $s$, the error is averaged over 100 networks
drawn under the corresponding SBMO. As expected, the algorithm performs
best with large values of $a$ (that correspond to larger
degrees). For each value of $a$, the case $s=1/2$ corresponds to a
standard SBM without overlap and we see that for $a$ large enough our algorithm performs
well. As $s$ decreases, we see that below a certain value of $s$ the
performance of the algorithm deteriorates greatly. There is a small rebund around $s=0.15$ which is simply due to the fact that for a purely random classification, the proportion of correctly classified nodes increases with the size of the overlap.

\begin{figure}[ht]
\centering
\includegraphics[height=0.12\textheight]{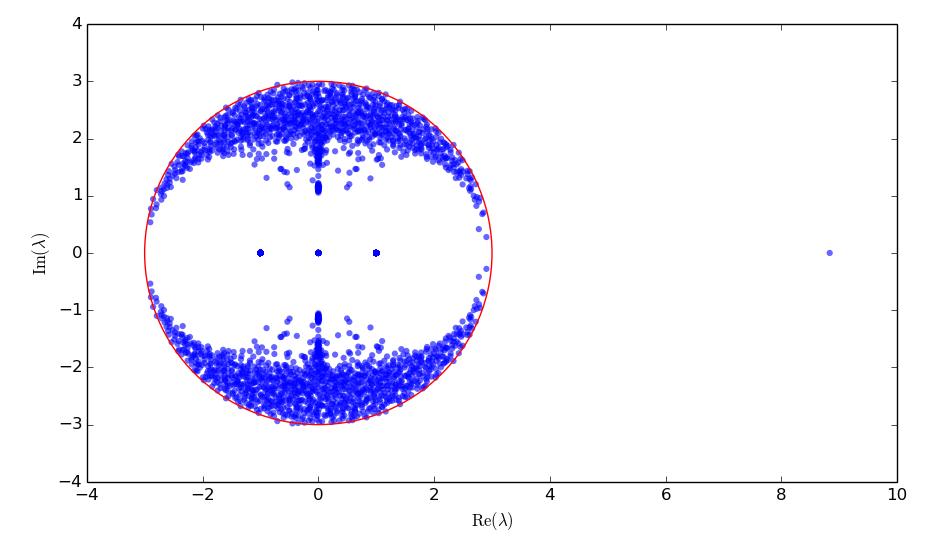}
\includegraphics[height=0.12\textheight]{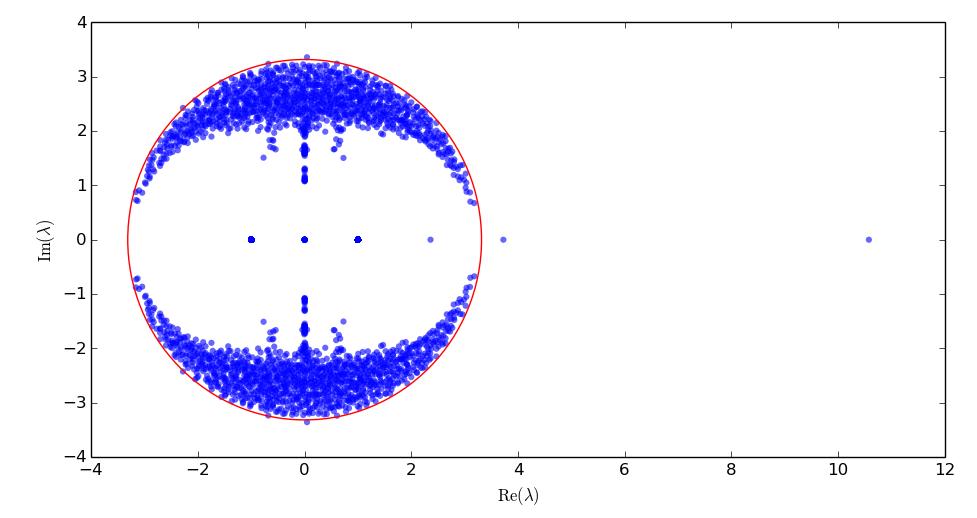}
\includegraphics[height=0.12\textheight]{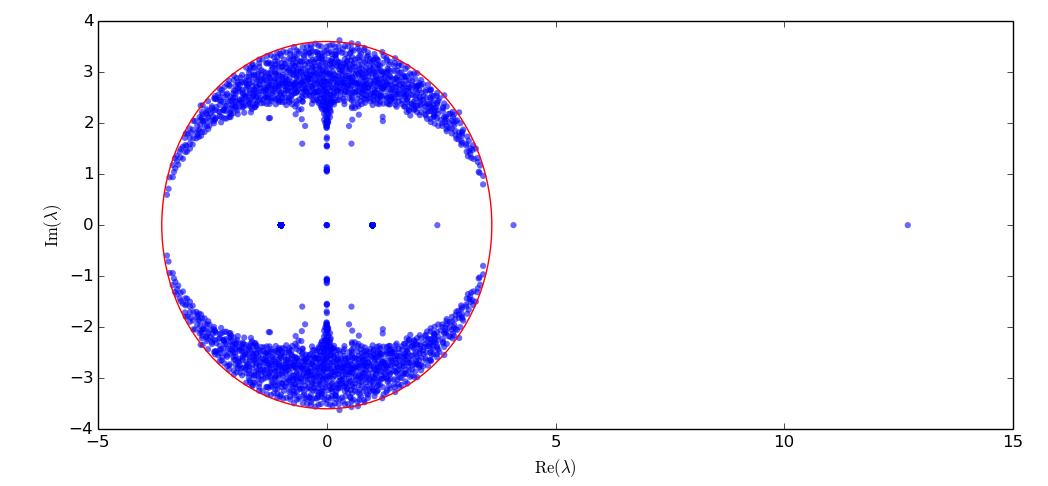}
\caption{\label{NBS} Spectrum of the non-backtracking operator with $n=1200$, $sn=400$ and $a=9,11,13$ and $b=0$. \\ The circle has radius $\sqrt{a(2-3s)}$ in each case.}
\end{figure}

We now investigate in more details possible improvements for our algorithm in the very sparse case.
A first conjecture may arise based on recent results obtained on the
non-backtracking matrix
\cite{krzakala2013spectral,bordenave2015non,saade2015spectral} which
can be seen as a way to regularize the adjacency matrix. 
The non-backtracking matrix is a representation of the link structure of a network that is an alternative to the usual adjacency matrix. 
A non-backtracking walk on a graph is a directed path such that no edge is the inverse of its preceding edge. The non-backtracking matrix of a graph is indexed by its directed edges and can be used to count non-backtracking walks of a given length. It has been used recently in the context of community detection and
we refer to
the works cited above for a precise description of the
non-backtracking matrix and its spectral analysis. We should stress
that to the best of our knowledge all results obtained so far for the non-backtracking
matrix requires the degrees in the graph to have the same average (which is not the case in the present framework). 
However, by analogy with the constant average degree case, the largest eigenvalue of the non-backtracking matrix
for our graph should be $(2-3s)(a+b) +o_n(1)$ and the noise, i.e. the
eigenvalues $\lambda$ corresponding to eigenvectors not correlated with
the communities should be of modulus $|\lambda|< \sqrt{(a+b)(2-3s)}$. 
Thus, if $s^2(a-b)^2>(2-3s)(a+b)$, a second eigenvalue appears
on the real axis at $s(a-b)+o_n(1)$. Hence we expect a spectral approach based on the 
non-backtracking matrix to be able to recover the overlapping communities whenever 
\begin{equation}
\frac{(a-b)^2}{a+b} \geq \frac{2-3s}{s^2}.\label{FirstConj} 
\end{equation}
In other words, a simple improvement of our adaptive SAAC would be to replace the selection of the eigenvectors of the adjacency matrix by the spectral procedure based on the non-backtracking matrix described in \cite{krzakala2013spectral}, in which case \eqref{FirstConj} would give the limit of this new algorithm. Note that detection is harder when $s$ is smaller, which is what happens for SAAC according to Figure~\ref{2B2overlap}. Figure \ref{NBS} illustrates the threshold \eqref{FirstConj}: the spectrum of the non-backtracking matrix is displayed for three values of $a$ (with $b=0$) around the phase transition that occurs when $a > \frac{2-3s}{s^2}$ which in this particular case is $9$ as $s=1/3$. 

\newpage

Recall that in the case of 2 symmetric communities in the SBM, the spectral method based on the non-backtracking matrix is known to be optimal \cite{bordenave2015non}. 
In our particular example of two overlapping communities, we believe that a naive application of the non-backtracking spectral approach will mot be optimal. We now describe a simple algorithm that should recover communities below the threshold \eqref{FirstConj} (at least for large values of $a$ and $b$). It can be checked that the average degree of a pure node in our model is $(1-s)(a+b)$ while the average degree of a node that belongs to both communities is $2(1-s)(a+b)$. 
Clustering (using, e.g. K-means) the degrees in two groups thus permits to identify the pure nodes from the mixed nodes (which may also be done using the first leading eigenvector of the adjacency matrix). Let pretend that we are able to remove all mixed nodes; then we consider the graph made of pure nodes only: it is drawn from a standard SBM and detection is possible (using the non-backtracking matrix) whenever 
\[\frac{(a-b)^2}{a+b} \geq 2,\]
which is smaller than \eqref{FirstConj}. After these two steps, the algorithm would have found the subcommunities and would need to build the communities from them.
This heuristic calculation (which might be made rigorous in a regime where $a,b\to \infty$ while $\frac{(a-b)^2}{a+b}\to c>2$ as done in \cite{caltagirone2016recovering}) suggests that the pre-processing step based on the degree clustering might help the spectral algorithm. Note however that this algorithm will detect subcommunities as described in Section \ref{sub:sub} and as explained there, its performance will degrade with an increasing number of communities.
We observe that this pre-processing step indeed also improves the performace of our algorithm. We compare this approach, called DC+SC (for Degree Clustering + Spectral Clustering) with SAAC on the examples of Figure~\ref{2B2overlap}.
We see that the performance is improved except in network with very few overlap in which the initial degree clustering is hard. Unfortunately, this algorithm seems difficult to extend to more generic SBMOs. Also, in very sparse network (like those of Section~\ref{sec:Experiments}) it would require to use the non-backtracking matrix in place of the adjacency matrix used in our simulated experiments.

\begin{figure}[ht]
\centering
\includegraphics[width=0.65\textwidth]{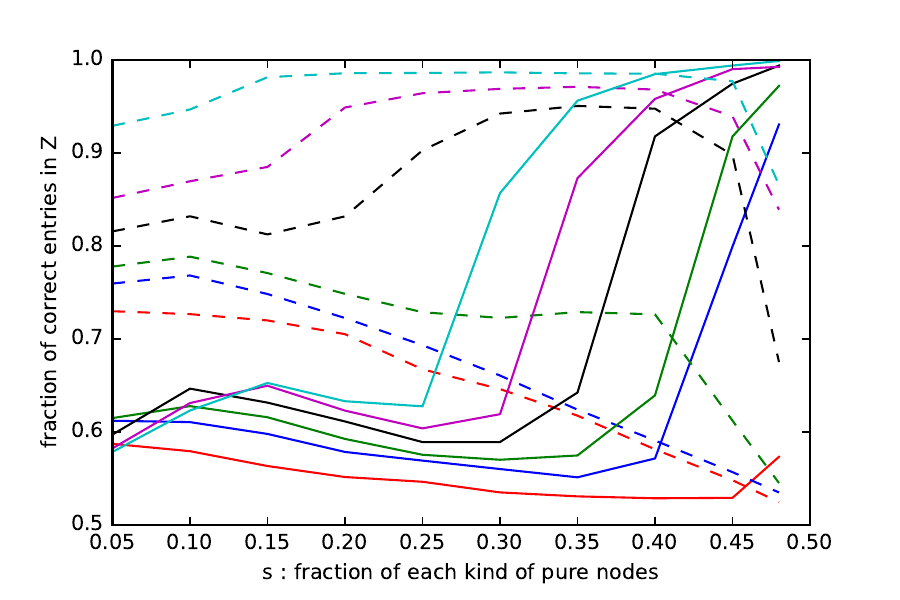}
\caption{\label{2B2overlapDashed} Performance of SAAC and the corresponding DC+SC algorithm (dashed line of the same color) \\  in the same setup as Figure~\ref{2B2overlap}}
\end{figure}

\section{Conclusion}

Most existing algorithms for community detection assume non overlapping communities.
Although they may in principle be used to detect all  {\it subcommunities}  generated by the various overlaps, this is not sufficient to recover the initial  communities due to the combinatorial complexity of the corresponding mapping. We have proposed a spectral algorithm, SAAC,  that works directly on the overlapping communities, using the specific geometry of the eigenvectors of the adjacency matrix under the SBMO. We have proved the consistency of this algorithm under the SBMO, provided each community has some positive fraction of pure nodes and the expected node degree is at least logarithmic, and tested its performance on both simulated and real data.

 This work has raised many interesting  issues. First, it would be worth relaxing the assumption that each community has some positive fraction of pure nodes, and have a spectral algorithm that is robust to extensions of the SBMO (including, e.g., degree corrections). Next, preliminary experiments on simulated data have shown threshold phenomena in the very sparse regime that should be further explored. Moreover we hope to understand how to use new sophisticated spectral tools based on the non-backtracking matrix for overlapping communities identification. 
Finally, the proof of consistency actually assumes that the underlying (NP-hard) optimization problem is solved exactly while this is not feasible in practice and heuristics need to be applied, like the proposed alternate optimization procedure. Understanding the impact of these heuristics on the performance of the algorithm is an interesting future research direction.

\paragraph{Acknowledgment} The authors acknowledge the support of the French Agence Nationale de la Recherche (ANR) under reference ANR-11-JS02-005-01 (GAP project).

\bibliography{biblioGraphs,biblioBandits}


\appendix

\section{Properties of the SBMO \label{proof:ID}\label{proof:FormEigen}}

\subsection{Identifiability: proof of Theorem~\ref{thm:ID}}

First note  that $A=ZBZ^T$ implies $\mathrm{rank}(A)\leq
\mathrm{rank}(B)$. Now condition (SBMO2) means that the restriction of $Z$ to its $K$ first rows is equal to $I_K$, up to some reordering of the nodes. This  gives  $\mathrm{rank}(A)\geq\mathrm{rank}(B)$, and thus $\mathrm{rank}(A)=\mathrm{rank}(B)$. If $B$ satisfies
(SBMO1), then  $\mathrm{rank}(A)=\mathrm{rank}(B)=K$:  the parameter $K$  is identifiable.

Now let  $Z,Z' \in \cZ$  and $B,B'$ invertible matrices such that 
$A=ZBZ^T = Z'B'Z'^T$.
We show that there exists some permutation  $\sigma\in  \mathfrak{S}_K$ such that $Z=Z'P_\sigma$ and $B=P_{\sigma}B'P_{\sigma}^T$.

 Let $U$ be a matrix containing $K$ independent normalized eigenvalues of $A$ associated to non-zero eigenvalues. The columns of $U$ form a basis of $\text{Im}(A)$. As $\text{Im}(A)\subset \text{Im}(Z)$ and $\text{Im}(A)\subset \text{Im}(Z')$, there exist invertible matrices $X, X'$ such that 
$U=ZX = Z'X'$.
As for all $k=1,\ldots,K$ there exists some $i$ such that $Z_{i,k}=\delta_{i,k}$, the $k$-th row of $X$ is a sum of rows in $X'$, namely
$$
X_k = \sum_{l \in \cS_k} X'_l, 
$$
where $\cS_k\subset \{1,\ldots,K\}$.
Similarly,  each row  of $X'$ is a sum of rows in $X$.
In particular, for any $k\ne l$, there exist $K$ integers $a_1,\ldots,a_K$ such that:
$$
X_k+X_l=\sum_{m=1}^{K}a_m X_m.
$$
If $\cS_k\cap\cS_l\neq \emptyset$,  there exists some $m$ such that $a_m\ge 2$. But this is in contradiction with the fact that $X$ is invertible.
Hence, $\cS_k\cap\cS_l =\emptyset$ for all $k\ne l$. The only way for the $\cS_k$ to be pairwise disjoint is that there exists a permutation $\sigma$ such that  $X'=P_{\sigma}X$. Since $ZX=Z'X'$ and $X$ is invertible, this implies 
$Z = Z'P_{\sigma}$. We deduce that 
$ZBZ^T = Z P_{\sigma^{-1}} B' P_{\sigma^{-1}}^T Z^T$ and
$B = P_{\sigma^{-1}} B'
P_{\sigma^{-1}}^T$,
 by the injectivity of $Z$.

\subsection{Identifiability for SBM: proof of Proposition~\ref{prop:ID}}

We simply prove that two nodes $i,j$ are in the same community if and only if 
$A_i=A_j$. This implies the identifiability of the model: it is indeed sufficient to group nodes whose  rows in $A$ are identical.
Let $i,j$ be such that  $A_i=A_j$.
If $Z_i\ne Z_j$ then $BZ_i^T\neq BZ_j^T$ by assumption (SBM1)
and $A_i =ZBZ_i^T\neq
ZBZ_j^T=A_j$ by assumption (SBM2), a contradiction.
Conversely, $Z_i=Z_j$ clearly implies $A_i=A_j$.

\subsection{Spectrum of the adjacency matrix}

\paragraph{Proof of Proposition~\ref{prop:FormEigen}}
As any non zero-eigenvector of $A$ belongs to $\text{Im}(A)\subseteq \text{Im}(Z)$, if $u$ is an eigenvector of $A$ associated to $\alpha_n\mu \neq 0$, there exists $x \in \R^K$ such that $u=Zx$. The following statements are equivalent:
\begin{eqnarray*}
 A (Zx) &=& \alpha_n \mu (Zx) \\
 \frac{\alpha_n}{n}ZBZ^TZx &=& \alpha_n \mu Zx \\
 Z B \left(\frac{1}{n}ZZ^T\right) x & = &\mu Z x \\
  B O x &= &\mu x \\
  BO^{1/2}(O^{1/2}x) = & = & \mu O^{-1/2}(O^{1/2}x) \\
  O^{1/2}BO^{1/2}(O^{1/2}x) & = & \mu(O^{1/2}x)
\end{eqnarray*}
Hence $Zx$ is an eigenvector of $A$ associated to $\alpha_n \mu$ if and only if $O^{1/2}x$ is an eigenvector of $O^{1/2}BO^{1/2}$ associated to $\mu$, which concludes the proof.

\section{Key ingredients in the proof of Theorem~\ref{thm:AnalysisASCO}} \label{sec:proofSensitivity}

\subsection{Proof of Lemma~\ref{prop:Decomposition}: Decomposition.}  

$\sqrt{n} U$ contains independent eigenvectors of $A$ associated to non-zero eigenvalues. From the first statement in Proposition~\ref{prop:FormEigen}, there exists a matrix $V$ of eigenvectors 
of $M_0$ such that $\sqrt{n}U = ZO^{-1/2} V.$ As $U$ contains normalized eigenvectors, $U^{T}U=I_K$, which yields $V^TV = I_K$ and $V\in\cO_K(\R)$.

\subsection{Proof of Lemma~\ref{prop:Robustness}: Sensitivity to noise}

Recall that from Proposition~\ref{prop:Decomposition}, there exists $V\in\cO_K(\R)$ such that the matrix of leading eigenvectors $U$ can be written  
\[U = ZX \ \ \ \text{with} \ \ \  X=\frac{1}{\sqrt{n}}ZO^{-1/2} V.\]
 Using that $||zX||=||zO^{-1/2}||/\sqrt{n}$, the following inequality is a consequence of the definition of $d_0$ (Definition~\ref{def:KeyQuantities}):
\begin{equation}\forall \ z \in \{-1,0,1,2\}^{1\times K} \backslash \{0\}, \ \ \ ||zX|| \geq \frac{d_0}{\sqrt{n}}.\label{Consequenced0}\end{equation}

Let $i_1,\dots,i_K$ (resp. $j_1,\dots,j_K$) be pure nodes in $\cN$ relatively to $Z$ (resp. $Z'$) that belong to communities $1,\dots K$:  
$Z_{i_k}=\ind_{\{k\}}$ (resp. $Z'_{j_k}=\ind_{\{k\}}$).
We first prove that $i_1,\dots,i_K$ are also pure nodes relatively to $Z'$. 
For any $k=1,\ldots,K$, ${Z}'_{i_k}$ can be written as a sum of pure nodes relatively to $Z$: there exists a set $\cS_k\subset \{1,\ldots,n\}$ such that 
\[{Z'}_{i_k} = \sum_{m\in \cS_k} {Z}_{j_m}\]
Let $k\neq l$. As $i_k$ and $i_l$ belong to $\cN$,  
\[||({Z}'_{i_k} + {Z}'_{i_l}) X' - (Z_{i_k} + Z_{i_l}) X || \leq ||{Z}'_{i_k}{X}' - Z_{i_k}X|| +  ||{Z}'_{i_l}{X}' - Z_{i_l}X||\leq \frac{d_0}{2\sqrt{n}}\]
and 
\begin{eqnarray*}
||({Z}'_{i_k} + {Z}'_{i_l}) {X}' - (\sum_{m\in \cS_k}\! Z_{j_m} + \sum_{m\in \cS_l}\! Z_{j_m}) X || &=  & ||(\!\sum_{m\in\cS_k}\!\!{Z}'_{j_m} + \sum_{m\in\cS_l}\!\!{Z}'_{j_m}\!) {X}' - (\!\sum_{m\in \cS_k}\!\! Z_{j_m} + \sum_{m\in \cS_l}\!\! Z_{j_m}\!) X || \\
&\leq& \sum_{m\in \cS_k} ||{Z}'_{j_m} {X}' - Z_{j_m}X || + \sum_{m\in \cS_l} ||{Z}'_{j_m} {X'} - Z_{j_m}X || \\
&\leq &\frac{d_0}{2\sqrt{n}}.
\end{eqnarray*}
This proves that 
\[||(\sum_{m\in \cS_k} Z_{j_m} + \sum_{m\in \cS_l} Z_{j_m})X - (Z_{i_k} + Z_{i_l}) X|| \leq {d_0}/{\sqrt{n}}. \]
If $\cS_k \cap \cS_l \neq \emptyset$, there exists $z\in \{0,1,2,-1\}\backslash \{0\}$ such that $||zX||\leq d_0/\sqrt{n}$, which contradicts~\eqref{Consequenced0}. Thus $\cS_k \cap \cS_l =\emptyset.$
Hence, the support of the ${Z}'_{i_k}$ are all disjoints, thus they must be distinct pure nodes. There exists a permutation $\sigma\in \mathfrak{S}_K$ such that 
\[\forall k=1,\ldots,K, \ \ Z_{i_k} = \ind_{\{k\}} \ \ \text{and} \ \ Z'_{i_k} = \ind_{\{\sigma(k)\}}.\]

To conclude the proof, we show that for $\sigma$ the permutation defined above, it holds that
\[\forall i \in \cN, \forall k\in \{1,K\}, \ \ \ {Z}'_{i,\sigma(k)} = Z_{i,k}.\]
Let $i\in\cN$. There exists a set $\cS\subset \{1,\ldots,n\}$ such that $Z_i = \sum_{k\in\cS}\ind_{\{k\}}$. It is sufficient to prove that $Z_i' = \sum_{k\in\cS}\ind_{\{\sigma(k)\}}$.
To do so, we first introduce $\cC = \{0,1\}^{1\times K} \backslash \{0\}$ and the following important mapping: 
\begin{eqnarray*}
\Phi : \ \  \cC & \longrightarrow & \cC\\
z & \longmapsto & y : z{X}' \in \cR_{y},
\end{eqnarray*} 
where $\R^K$ is partitioned into the following $2^K-1$ regions indexed by $y\in \cC$, 
\[\cR_y = \{ x \in \R^{1\times K} : ||x - yX|| < ||x - y'X|| \ \text{for all} \ \ y'\in \cC, y'\neq z\}. \]
The following lemma gathers useful properties of the mapping $\Phi$. Its proof is given below.

\begin{lemma}\label{lem:PropPhi}  $\Phi$ is a one-to-one mapping satisfying 
$||zX' - yX|| \leq \frac{d_0}{2\sqrt{n}} \ \Rightarrow \ \Phi(z)=y.$
\end{lemma}

As $i\in\cN$, from assumption 1., 
\[||Z_i'X' - Z_iX|| \leq \frac{d_0}{2\sqrt{n}}.\]
Moreover, using that $Z_i = \sum_{k\in\cS}\ind_{\{k\}} = \sum_{k\in\cS} Z_{i_k}$ and $\sum_{k\in\cS}\ind_{\{\sigma(k)\}} = \sum_{k\in\cS} Z'_{i_k}$, one has 
\begin{eqnarray*}
\left|\left|\left(\sum_{k\in\cS} \ind_{\sigma(k)}\right){X}' - Z_iX \right|\right| & = &  \left|\left|\left(\sum_{k\in\cS} {Z}'_{i_k}\right){X}' - \left(\sum_{k\in \cS} Z_{i_k}\right) X \right|\right| \\
& \leq & \sum_{k\in\cS} \left|\left|{Z}'_{i_k}{X'} -  Z_{i_k} X \right|\right| \leq \frac{d_0}{2\sqrt{n}}.
\end{eqnarray*}
Using Lemma~\ref{lem:PropPhi}, the last two inequalities yield $\Phi(Z_i')=Z_i$ and $\Phi\left(\sum_{k\in\cS} \ind_{\sigma(k)}\right)=Z_i$ respectively. Using that $\Phi$ is one-to-one (again from Lemma~\ref{lem:PropPhi}) concludes the proof: 
\[Z_i' =\sum_{k\in\cS} \ind_{\sigma(k)}.\]

\paragraph{Proof of Lemma~\ref{lem:PropPhi}} Let $z,y\in\cC$ be such that $||zX' - yX|| \leq \frac{d_0}{2\sqrt{n}}$. Let $y'\in \cC$ : $y'\neq y$. Using~\eqref{Consequenced0},  
\[|| {z}{X'} - y'X || > || z X-y'X|| - || {z}X' - yM|| \geq \frac{d_0}{\sqrt{n}} - \frac{d_0}{2\sqrt{n}}  > \frac{d_0}{2\sqrt{n}} \geq ||{z}{X'} - yX||.\]
Hence, $zX' \in \cR_{y}$ and $\Phi(z)=y$, which proves the second part of the result. 

We now prove that $\Phi$ is one-to-one. Let $y\in \cC$: there exists a set $\cS \subseteq \{1,\dots,K\}$ such that $y=\sum_{m\in \cS}\ind_{\{k\}}=\sum_{m\in\cS} Z_{i_m}.$
Let $z=\sum_{m\in \cS} {Z}'_{i_m}$. As the ${Z}'_{i_m}$ are disjoint indicators, one has $z\in \cC$. Moreover, 
\[||z{X}' - yX || \leq \sum_{m\in\cS} \left|\left|{Z}'_{i_m}X' - Z_{i_m}X \right|\right| \leq \frac{d_0}{2\sqrt{n}}.\]
From what we've just proved, this implies $\Phi(z)=y$. As $\cC$ is finite and $\forall y\in\cC,\exists z\in\cC : \Phi(z)=y$, $\Phi$ is one-to-one.

\section{Results for non-adaptive procedures \label{sec:NonAdaptive}}

We present here tighter upper bounds on the fraction of nodes that are misclassified by some non-adaptive estimation procedures, based on $\hat{U}\in \R^{n\times K}$ rather than on $\hat{U}\in \R^{n\times \hat{K}}$ (with $\hat{K}$ given in Theorem~\ref{thm:AnalysisASCO}). In this case, it is possible to analyze the solution of $(\cP_\epsilon)$, defined in Section~\ref{subsec:Consistency}, as well as the solution of the following optimization problem: 
\[
 \left(\cP_{\cT}\right) : \ \ \underset{\substack{Z'\in \{0,1\}^{n,K} : \forall i, Z'_i \in \cT \\X' \in \R^{K\times K}}}{\min} \ ||Z'X' - \hat{U} ||_{F}^2.
\] 
$(\cP_{\cT})$ relies on the knowledge of $\cT$, the set of subcommunities that are present in the network. If one has this knowledge, note that the above estimate can be computed using alternate minimization, just like the solution of $(\cP)'$. Theorem~\ref{thm:AnalysisASCONonAdaptive} below gathers the theoretical guarantees obtained for these two estimators. Compared to Theorem~\ref{thm:AnalysisASCO}, a logarithmic factor is removed in the upper bound on the number of misclassified nodes: both estimates are consistent provided that $\alpha_n \geq (L_{\max}^{-1})\log(n)$.

\begin{theorem}\label{thm:AnalysisASCONonAdaptive} Let $\hat{U}$ be a matrix formed by ${K}$ independent eigenvectors associated to the eigenvalues of $\hat{A}$ that are largest in absolute value. Let
$(\hat{Z},\hat{C})$ be the solution of $(\cP_{\epsilon})$ or of $(\cP_{\cT})$. 

For all $r > 0$, there exists a constant $C_r$ such that if $\alpha_n \geq (L_{\max}^{-1})\log(n)$
then, for $n$ large enough, with probability larger than $1-n^{-r}$, 
\[
\frac{\MisC(\hat{Z},Z)}{n} \leq C_r\frac{K^3L_{\max}}{d_0^2\mu_0^2}\frac{1}{\alpha_n}.
\]
\end{theorem}

The proof of Theorem~\ref{thm:AnalysisASCONonAdaptive} is very similar to that of Theorem~\ref{thm:AnalysisASCO} given in the previous section. The main difference is that in the non-adaptive case it is possible to use a tighter eigenvectors perturbation result (specific to SBMO), that we state below as Lemma~\ref{lem:FroebeniusAdj}. Compared to Lemma~\ref{lem:FroebeniusAdjEmpOptim}, in  Lemma~\ref{lem:FroebeniusAdj} an extra logarithmic factor is removed, but at the price of non-explicit constants, that do not permit to propose an adaptive version of the result. The proof of both Lemma~\ref{lem:FroebeniusAdjEmpOptim} and Lemma~\ref{lem:FroebeniusAdj} are given in the next section. 

\begin{lemma}\label{lem:FroebeniusAdj} Let $\hat{A}$ be drawn under a SBMO model with expected adjacency matrix $A$. 
Let $K$ be the rank of ${A}$. Let ${U}$ (resp. $\hat{U}$) be a matrix whose columns are $K$ independent eigenvectors associated to the $K$ eigenvalues of ${A}$ (resp. $\hat{A}$) with largest absolute values.  

For all $r>0$, there exists a constant $C_r$ such that if  $\dmax \geq \log \left(n\right)$,
with probability larger than $1-n^{-r}$, there exists a matrix $\hat{P}\in \cO_n(\R)$ such that  
\[\left|\left|\hat{U} - {U}\hat{P}\right|\right|_F^2 \leq KC_r\left(\frac{\dmax}{\lambda_{\min}({A})^2}\right).\]
\end{lemma}

Also, compared to that of $(\cP_\epsilon)$, the analysis of the solution of $(\cP_{\cT})$ requires a more complex argument to prove that the set $\cN_n$ and ($\hat{Z},\hat{X_1})$ defined in the proof of Theorem~\ref{thm:AnalysisASCO} satisfy assumption 2. of Lemma~\ref{prop:Robustness}, i.e. that $\cN_n$ contains one pure nodes per community in $Z$ and $\hat{Z}$. We present below the argument that can be used in that case.

\paragraph{$\cN_n$ contains pure nodes} 
Under the assumption 
\[
\frac{64K^2}{d_0^2}||\hat{U} - {U}\hat{P}||_F^2 \leq \min_{z \in \cT} \ \beta_z,
\]
${|\cN_n^c|}/{n}\leq \beta_z$ for each possible membership vector $z\in \cT$. Thus, for all $z\in\cT$, the set of nodes $i$ such that $Z_i = z$ cannot be included in $\cN_n^c$ and there exists $i_z\in\cN_n$ such that $Z_{i_z} = z$. In particular, $\cN_n$ contains pure nodes relatively to $Z$. Now we need to prove that it also contains pure nodes relatively to $\hat{Z}$. 

To do so, we introduce the following mapping and prove it is one-to-one: 
\begin{eqnarray*}
\Psi : \ \  \cT & \longrightarrow & \cT\\
z & \longmapsto & y : z\hat{X}_1 \in \cR_{y},
\end{eqnarray*}
where $\R^K$ is partitioned into $\tilde{K}=|\cT|$ regions, indexed by $y\in \cT$, 
\[{\cR}_y = \{ x \in \R^K : ||x - yX|| < ||x - y'X|| \ \text{for all} \ \ y'\in \cT : y'\neq y\}. \]
For all $i\in\cN_n$, $\Psi(\hat{Z}_i)=Z_i$. Indeed, $\hat{Z}_i \hat{X}_1 \in \tilde{\cR}_{Z_i}$ for if $y'\in\cT$ is such that $y'\neq Z_i$, using \eqref{Consequenced0} and the fact that $i$ belongs to $\cN_n$ yields
\[|| \hat{Z}_i\hat{X}_1 - y'X || > || {Z}_i X-y'X|| - || \hat{Z}_i\hat{X}_1 - Z_iX|| \geq \frac{d_0}{\sqrt{n}} - \frac{d_0}{2\sqrt{n}}  > \frac{d_0}{2\sqrt{n}} \geq || \hat{Z}_i\hat{X}_1 - Z_iX||.\]
It follows that for all $y\in\cT$, there exists $z\in\cT$ such that $y=\Psi(z)$. Indeed, there exists $i_y\in\cN_n$ such that $Z_{i_{y}} = y$, thus $\Psi(\hat{Z}_{i_y})=y$ and $z=\hat{Z}_{i_y}$ belongs to $\cT$ by definition of the optimization problem that $\hat{Z}$ solves. As $\cT$ is a finite set, $\Psi$ is one-to-one. Thus, one has
\[\left\{\hat{Z}_{i_{z}} : z\in \cT\right\} = \Psi^{-1}\left(\left\{{Z}_{i_{z}} : z\in \cT\right\}\right) = \Psi^{-1}\left(\cT\right) = \cT.\]
In particular, there exists  $i_1,\dots,i_K$ (resp. $j_1,\dots,j_K$) such that $\forall k \in \{1,\dots,K\}$, $Z_{i_k}=\hat{Z}_{j_k}=\ind_{\{k\}}$.

\section{Proof of the eigenvectors perturbation results\label{proof:Estimation}}

Lemma~\ref{lem:FroebeniusAdjEmpOptim} and Lemma~\ref{lem:FroebeniusAdj} rely on two main ingredients, that we now introduce. First, a (deterministic) eigenvector perturbation result, extracted from \cite{LeiRinaldo15} and second a new high-probability upper bound on the spectral norm of $\hat{A}-A$. 

\subsection{Main ingredients}

\paragraph{Algebraic results} 

We rephrase below Lemma 5.1 in \cite{LeiRinaldo15} that relates the Froebenius distance between the matrices of leading eigenvectors of two matrices in terms to their distance in spectral norm. This important result relies on algebraic tools, notably the Davis-Kahan theorem \cite{DavisKahan70}. 

\begin{lemma}( Lemma 5.1 in \cite{LeiRinaldo15})\label{lem:AlgRinaldo} Let $A$ and $B$ be two $n\times n$ symmetric matrices such that $A$ has rank $K$. Let $X_A$ (resp. $X_B$) be a matrix of orthonormal eigenvectors associated to the eigenvalues with largest absolute values of $A$ (resp. $B$). There exists a $K\times K$  orthogonal matrix $P \in \cO_{K}(\R)$ such that 
\[ ||X_B - X_A P ||_{F} \leq \frac{2\sqrt{2K}}{\lambda_{\min}(A)}||A - B|| \]
\end{lemma}

In the proof of Lemma~\ref{lem:FroebeniusAdjEmpOptim}, we shall also use directly the closeness of the leading eigenvalues of $A$ and $\hat{A}$. For this purpose, we state here a consequence of the Weyl inequalities. 

\begin{lemma}[Weyl's inequalities]\label{lem:Weyl} Let $\lambda_1(M) \geq \dots \geq \lambda_n(M)$ denote the ordered eigenvalues of an $n\times n$ symmetric matrix $M$. For any two symmetric matrices $A$ et $B$ of size $n$,
\[\text{for all} \ \ i=1,\dots,n \ \ |\lambda_i(A) - {\lambda}_i(B)| \leq ||A - B||.  \] 
\end{lemma}

\paragraph{Deviation inequalities}

Using Lemma~\ref{lem:AlgRinaldo} above, one may express the distance between $\hat{U}$ and $U$ as a function of the spectral norm $||\hat{A} - A||$. The next step is thus to control this quantity, which can be done using some matrix concentration inequality, namely a Bernstein inequality for the sum of independent matrices.  We state below our new deviation result, that is of interest in its own and holds in any random graph model. The proof is postponed to \ref{proof:ConcPerso}.

\begin{theorem}\label{thm:ConcPerso} Let $\delta \in ]0,1[$. Let $\epsilon>0$ be fixed. If  
\[\dmax \geq \frac{2}{9}\frac{1+\epsilon}{\epsilon^2}\log \frac{2n}{\delta},\]
one has 
\[\bP\left(||\hat{A} - A || > \sqrt{2(1+\epsilon)\dmax\log\left(\frac{2n}{\delta}\right)}\right)\leq   \delta.\]
\end{theorem}

Another concentration result, given below, is used to prove Lemma~\ref{lem:FroebeniusAdj}. This result, recently obtained by \cite{LeiRinaldo15} improves the dependency in $n$ in the high-probability upper bound on $||\hat{A}-A||$, since a logarithmic term is removed compared to Theorem~\ref{thm:ConcPerso}. However, the constants in the upper bound are non-explicit. 

\begin{theorem}\label{thm:ConcLeiRinaldo}[Theorem 5.2 of \cite{LeiRinaldo15}] In a random graph model, if $d$ is such that 
$d \geq n \max_{i,j} A_{i,j}$ and $d \geq c_0 \log(n)$, for every $r>0$ there exists a constant $C=C(r,c_0)$ such that 
 \[\bP\left(||\hat{A} - A || > C\sqrt{d}\right)\leq   \delta.\]
\end{theorem}

Finally the proof of Lemma~\ref{lem:FroebeniusAdjEmpOptim} also requires another result to control the deviations of the empirical degrees from the mean degrees. Lemma~\ref{lem:DegreesFirst} follows from Bernstein inequalities for independent random variables and is proved in~\ref{proofs:Degrees}. 

\begin{lemma}\label{lem:DegreesFirst} Let $\alpha \in ]0,1[$.
\begin{eqnarray*}
\bP\left(\dmaxE \leq (1+\alpha)\dmax\right) &\geq& 1 - ne^{-\dmax \frac{\alpha^2}{2(1+\alpha/3)}}  \\
\bP\left(\dmaxE \geq (1-\alpha)\dmax\right) &\geq& 1 - e^{-\dmax \frac{\alpha^2}{2(1+\alpha/3)}} 
\end{eqnarray*}
\end{lemma}

\subsection{Proof of Lemma \ref{lem:FroebeniusAdjEmpOptim} and Lemma~\ref{lem:FroebeniusAdj}}

Let $U_K$ (resp. $\hat{U}_K$) be a matrix whose columns are $K$ orthogonal eigenvectors associated to the largest eigenvalues (in absolute value) of matrix $A$ (resp. $\hat{A}$). Applying Lemma~\ref{lem:AlgRinaldo} to the expected adjacency matrix $A$, that has rank $K$, and to the matrix $B=\hat{A}$ yields that there exists $\hat{P}\in \cO_K(\R)$ such that the following inequality holds:
\begin{equation}
||\hat{U}_K - {U}_K\hat{P}||_F^2 \leq \frac{8K}{\lambda_{\min}({A})^2}||\hat{A}-{A}||^2.
\label{ImportantThing}
\end{equation}

\paragraph{Proof of Lemma~\ref{lem:FroebeniusAdjEmpOptim}}
Let $\eta$ be fixed and let $\epsilon=\eta/(2+\eta)$, so that $(1+\epsilon)/(1-\epsilon)=1+\eta$. Let $\cE,\cF,\cG$ be the three events
\begin{eqnarray*}
 \cE &=& \left(||\hat{A} - A || \leq \sqrt{2(1+\epsilon)\dmax\log(4n/\delta)}\right)  \\
 \cF &=& \Big(\hat{d}_{\max} \leq (1+\epsilon)d_{\max}\Big)\\
 \cG & =& \Big(\hat{d}_{\max} \geq (1-\epsilon)d_{\max}\Big)
\end{eqnarray*}
and $\cH = \cE \cap \cF \cap \cG$. We first show that $\bP(\cH) \geq  1-\delta$ under the assumption 
\begin{equation}
\dmax \geq \frac{2(1+\epsilon/3)}{\epsilon^2} \log \left(\frac{4n}{\delta}\right).
\label{assum1}
\end{equation}
From Theorem~\ref{thm:ConcPerso}, this condition implies $\bP(\cE^c) \leq \delta/2$. From Lemma~\ref{lem:DegreesFirst}, one has
\begin{eqnarray*}
 \bP(\cF^c) &\leq &ne^{-\dmax \frac{\epsilon^2}{2(1+\epsilon/3)}} \leq {\delta}/{4}, \\
 \bP(\cG^c) &\leq& e^{-\dmax\frac{\epsilon^2}{2(1+\epsilon/3)}}  \leq ne^{-\dmax\frac{\epsilon^2}{2(1+\epsilon/3)}}  \leq {\delta}/{4}.
\end{eqnarray*}
A union bound then yields $\bP(\cH)\geq 1-\delta$.

We now assume that the event $\cH$ holds. Recall $\lambda_k(A)$ (resp. $\lambda_k(\hat{A})$) are the eigenvalues of $\hat{A}$ (resp. $\hat{A}$)  sorted in non-increasing order. We first prove that under the extra assumption 
\begin{equation}
\lambda_{\min}(A) \geq C_\epsilon\sqrt{\dmax\log(4n/\delta)} \ \ \ \text{with} \ \ \ C_\epsilon = \sqrt{2(1+\epsilon)}\left(1+\sqrt{\frac{1+\epsilon}{1-\epsilon}}\right),
\label{assum2}
\end{equation}
the set 
\[\hat{S}_n^{\epsilon} = \left\{ k : |\lambda_k(\hat{A})| > \sqrt{2\frac{1+\epsilon}{1-\epsilon}\dmaxE \log(4n/\delta)}\right\}\]
coincides with the set of $K$ leading eigenvalues of $A$, and is, in particular, of cardinality $K$.   

 Let $s$ (resp. $r$) be the number of of eigenvalues of ${A}$ that are strictly positive (resp. negative), so that the rank of $A$ satisfies $K=s+r$. We show that $\hat{S}_n^{\epsilon}$ coincides with $\{1,s\}\cup\{n-r+1,n\}$. Using Weyl's inequalities (Lemma~\ref{lem:Weyl}), one can write
\begin{eqnarray*}
\text{for} \ k= 1,\dots s, \ \ \ \ \lambda_k(\hat{A}) & \geq & \lambda_k({A}) - ||\hat{A}-{A}||, \\
\text{for} \ k=s+1,\dots,n-r, \ \ \ \ |\lambda_k(\hat{A})| & \leq & ||\hat{A}-{A}||, \\
\text{for} \ k=n-r+1,\dots,n, \ \ \ \ \lambda_k(\hat{A}) & \leq & \lambda_k({A}) + ||\hat{A}-{A}||. 
\end{eqnarray*}
As a consequence, using that event $\cE$ holds, one has  
\begin{eqnarray}
\text{for} \ k \notin\{s+1,n-r\}, \ \ \ \ |\lambda_k(\hat{A})| & > & {\lambda_{\min}(A)} - \sqrt{2(1+\epsilon)d_{\max} \log(4n/\delta)} \label{ToUse1}\\
\text{for} \ k \in \{s+1,n-r\} \ \ \ \ |\lambda_k(\hat{A})| & < & \sqrt{2(1+\epsilon)\dmax \log(4n/\delta)}. \label{ToUse2} 
\end{eqnarray}
For every $k\in\hat{S}_n^{\epsilon}$, using that $\cG$ holds, one has 
\begin{eqnarray*}
|\lambda_k(\hat{A})| &>& \sqrt{2\frac{1+\epsilon}{1-\epsilon}\dmaxE \log(4n/\delta)}\geq \sqrt{2(1+\epsilon) d_{\max} \log(4n/\delta)}.
\end{eqnarray*}
From Inequality \eqref{ToUse2}, this proves that $k \in \{1,n\} \backslash \{s+1,n-r\}$. Conversely, Let $k\in \{1,n\}\backslash\{s+1,n-r\}$. Using Inequality \eqref{ToUse1},
\begin{eqnarray*}
|\lambda_k(\hat{A})| &\geq& C_\epsilon\sqrt{\dmax\log(4n/\delta)} - \sqrt{2(1+\epsilon)\dmax\log(4n/\delta)}  \\
& \geq & (C_\epsilon - \sqrt{2(1+\epsilon)}) \sqrt{(\dmaxE/(1+\epsilon)) \log(4n/\delta)}  = \frac{C_\epsilon-\sqrt{2(1+\epsilon)}}{\sqrt{1+\epsilon}}\sqrt{\dmaxE \log(4n/\delta)} \\
& > & \sqrt{2\frac{1+\epsilon}{1-\epsilon}\dmaxE \log(4n/\delta)},
\end{eqnarray*}
where we use that $\cF$ holds for the second inequality. Hence $k\in \hat{S}_n^{\epsilon}$. Thus $\hat{S}_n^{\epsilon}=\{1,n\}\backslash \{s+1,n-r\}$. 

As the set $\hat{S}_n^{\epsilon}$ is of cardinality $K$, the matrix $\hat{U}$ in the statement of Lemma~\ref{lem:FroebeniusAdjEmpOptim} coincides with $\hat{U}_K$, the matrix formed by the $K$ leading eigenvectors of $\hat{A}$. From \eqref{ImportantThing}, there exists $\hat{P}$ such that
\[||\hat{U} - U\hat{P} || = ||\hat{U}_K - U_K\hat{P} || \leq \frac{8K}{\lambda_{\min}(A)^2}||\hat{A} - A||^2 \leq \frac{16(1+\epsilon)K}{\lambda_{\min}(A)^2}\dmax \log \left(\frac{4n}{\delta}\right).\]
The result follow by substituting $\epsilon$ with $\eta$ in this last equation and in assumptions \eqref{assum1} and \eqref{assum2}.

\paragraph{Proof of Lemma~\ref{lem:FroebeniusAdj}} In the SBMO model, there exists a constant $c$ such that 
\[\dmax \leq n \max_{i,j} A_{i,j} \leq c \dmax.\]
Let $r>0$. From Theorem~\ref{thm:ConcLeiRinaldo}, there exists a constant $\tilde{C}_r$ such that if $\dmax \geq \log(n)$, with probability larger than $1-n^{-r}$, 
\[||\hat{A} - A || \leq \tilde{C}_r\sqrt{n \max_{i,j} A_{i,j}} \leq (\sqrt{c}\tilde{C}_r)\sqrt{\dmax}.\]
From \eqref{ImportantThing}, this yields 
\[||\hat{U}_K - {U}_K||_F^2 \leq \frac{8K}{\lambda_{\min}({A})^2}||\hat{A}-{A}||^2 \leq \frac{8Kc\tilde{C}_r^2}{\lambda_{\min}({A})^2}\dmax = \frac{C_r K\dmax}{\lambda_{\min}(A)^2},\]
letting $C_r = 8 c \tilde{C}_r^2$.

\subsection{Proof of Theorem~\ref{thm:ConcPerso}: a matrix concentration result\label{proof:ConcPerso}}  

Our proof is based on the following result by \cite{Tropp12}.

\begin{lemma}[Theorem 1.4, \cite{Tropp12}]\label{lem:Tropp} Let $(X_k)$ be a sequence of independent, random, symmetric matrices with dimension $d$. Assume that each random matrix satisfies 
\[\bE[X_k] = 0 \ \ \text{and} \ \ \ \lambda_{\text{max}}(X_k) \leq R \ \ \text{almost surely}\]
and let $\sigma^2$ be such that $||\sum_{k=1}^n \bE[X_k^2] || \leq \sigma^2$. Then, for all $t\geq 0$, 
\[\bP\left(\lambda_{\max}\left(\sum_{k=1}^n X_k\right) \geq t\right) \leq d \exp\left(-\frac{t^2}{2(\sigma^2 + Rt/3)}\right).\] 
\end{lemma}

One has
\begin{eqnarray*}
 \hat{A} - A=\sum_{i\leq j} X_{i,j}, 
\end{eqnarray*}
where $X_{i,j}$ is a matrix of size $n$ defined by
\[X_{i,j} : = (\hat{A}_{i,j} - A_{i,j})\times
\left\{
\begin{array}{lcl}
 e_ie_j^T + e_je_i^T \ \ &  \text{if} \ \ &  i<j \\
 e_ie_i^T & \text{if} & i=j. \\
\end{array}
\right.
\]
One has $||X_{i,j}|| \leq |\hat{A}_{i,j} - A_{i,j}| \leq 1$ and 
\[\left|\left|\sum_{i\leq j} \bE[X_{i,j}^2] \right|\right|= \left|\left|\text{Diag}_i\left(\sum_{j=1}^n \bE[X_{i,j}^2]\right)\right|\right| = \max_i\sum_{j=1}^n {A}_{i,j}(1-A_{i,j})\leq  \max_i\sum_{j=1}^n {A}_{i,j} \leq \dmax.\]
From Lemma~\ref{lem:Tropp}, 
\[\bP\left(||\hat{A} - A || > \alpha \dmax\right)\leq 2n\exp\left(-\dmax\frac{\alpha^2}{2(1+\alpha/3)}\right)\]
Let $\epsilon>0$. Choosing $\alpha = \sqrt{2(1+\epsilon)\log(2n/\delta)/\dmax}$, for
\[\dmax \geq \frac{2}{9}\frac{1+\epsilon}{\epsilon^2}\log \frac{2n}{\delta}\]
(which is equivalent to $\alpha/3\leq \epsilon$), one has 
\[\bP\left(||\hat{A} - A || > \sqrt{2(1+\epsilon)\dmax\log\left(\frac{2n}{\delta}\right)}\right)\leq  2n\exp\left(- \frac{2(1+\epsilon)\log(2n/\delta)}{2(1 + \alpha/3)}\right) \leq \delta.\]

\subsection{Proof of Lemma \ref{lem:DegreesFirst}: a deviation result for the empirical degrees\label{proofs:Degrees}}

For all $i\in\{1,n\}$,  
\[\hat{d}_i - d_i = \sum_{j=1}^n (\hat{A}_{i,j} - A_{i,j}).\]
As $\bE[\hat{A}_{i,j}]=A_{i,j}$, $\hat{A}_{i,j}\leq 1$ and  $\sum_{j=1}^n \bE[\hat{A}_{i,j}^2] = \sum_{j=1}^n \bE[A_{i,j}] = d_i$, Bennett's inequality (see, e.g., Theorem 2.9 in \cite{Boucheronal13CI}) yields, for all $t>0$
\begin{eqnarray*}
\bP\left(\hat{d}_i - d_i  > t\right) &\leq& \exp\left(-d_i h\left(\frac{t}{d_i}\right)\right),\\
\bP\left(\hat{d}_i - d_i  < -t\right) &\leq& \exp\left(-d_i h\left(\frac{t}{d_i}\right)\right),
\end{eqnarray*}
where $h$ is the function defined by $h(u)=(u+1)\log(u+1)-u$.

Using the fact that $v\mapsto vh(t/v)$ is decreasing for all $t$, one obtains
\begin{eqnarray}
\bP\left(\hat{d}_i - d_i  > \alpha d_{\max}\right) &\leq& \exp\left(-d_{\max} h\left({\alpha}\right)\right),\label{Eq1}\\
\bP\left(\hat{d}_i - d_i  < -\alpha d_{\max}\right) &\leq& \exp\left(-d_{\max} h\left({\alpha}\right)\right)\label{Eq2}.
\end{eqnarray}
Let $i_0$ be such that $d_{i_0}=d_{\max}$. From \eqref{Eq2},
\begin{eqnarray*}
\bP(\hat{d}_{i_0} \geq d_{i_0}-\alpha d_{\max}) & \geq & 1 - e^{-d_{\max} h(\alpha)} \\
\bP(\hat{d}_{\max} \geq (1-\alpha)d_{\max}) & \geq & 1 - e^{-d_{\max} h(\alpha)},
\end{eqnarray*}
using that  in particular $\hat{d}_{\max} \geq \hat{d}_{i_0}$.
From \eqref{Eq1} and a union bound, 
\begin{eqnarray*}
\bP\left(\forall i\in\{1,n\}, \hat{d}_i \leq d_i + \alpha d_{\max}\right) &\geq& 1-n\exp\left(-d_{\max} h\left({\alpha}\right)\right), \\
\bP\left(\forall i\in\{1,n\}, \hat{d}_i \leq (1+\alpha)d_{\max}\right) &\geq& 1-n\exp\left(-d_{\max} h\left({\alpha}\right)\right), \\
\bP\left(\dmaxE \leq (1+\alpha)d_{\max}\right) &\geq& 1-n\exp\left(-d_{\max} h\left({\alpha}\right)\right),
\end{eqnarray*}
by definition of $\dmaxE$.
The statements in Lemma~\ref{lem:DegreesFirst} follow from the lower bound 
\[h(u) \geq \frac{u^2}{2(1+u/3)},\]
that can be found in \cite{Boucheronal13CI}.

\end{document}